\documentclass[10pt,twocolumn,letterpaper]{article}

\usepackage[pagenumbers]{cvpr}      

\usepackage[utf8]{inputenc} 
\usepackage[T1]{fontenc}    
\usepackage{url}            
\usepackage{booktabs}       
\usepackage{amsfonts}       
\usepackage{nicefrac}       
\usepackage{microtype}      
\usepackage{xcolor}         
\usepackage{svg}

\usepackage{soul}
\usepackage{graphicx}
\usepackage{amsmath,bm,amssymb}
\usepackage{multirow}
\usepackage{subcaption}
\usepackage{caption}

\usepackage[linesnumbered,ruled,vlined]{algorithm2e}

\definecolor{cvprblue}{rgb}{0.21,0.49,0.74}
\usepackage[pagebackref,breaklinks,colorlinks,allcolors=cvprblue]{hyperref}

\def\paperID{*****} 
\def\confName{CVPR}
\def\confYear{2026}

\title{Delta Sampling: Data-Free Knowledge Transfer
Across Diffusion Models}

\author{First Author\\
Institution1\\
Institution1 address\\
{\tt\small firstauthor@i1.org}
\and
Second Author\\
Institution2\\
First line of institution2 address\\
{\tt\small secondauthor@i2.org}
}

\author{
Zhidong Gao\\
Shanxi University\\
\texttt{zhidong.gao@sxu.edu.cn} \\
\and
Zimeng Pan\\
Google Cloud\\
\texttt{justinpan@google.com}\\
\and 
Yuhang Yao\\
Carnegie Mellon University\\
\texttt{yuhangya@alumni.cmu.edu}\\
\and 
Chenyue Xie\\
University of Science and Technology of China\\
\texttt{cyxie@ustc.edu.cn}\\
\and 
Wei Wei$^\dagger$\\
Shanxi University \\
\texttt{weiwei@sxu.edu.cn}\\
}

\begin{document}
\maketitle
\begin{abstract}
Diffusion models like Stable Diffusion (SD) drive a vibrant open-source ecosystem including fully fine-tuned checkpoints and parameter-efficient adapters such as LoRA, LyCORIS, and ControlNet. However, these adaptation components are tightly coupled to a specific base model, making them difficult to reuse when the base model is upgraded (e.g., from SD 1.x to 2.x) due to substantial changes in model parameters and architecture. In this work, we propose Delta Sampling (DS), a novel method that enables knowledge transfer across base models with different architectures, without requiring access to the original training data. DS operates entirely at inference time by leveraging the delta: the difference in model predictions before and after the adaptation of a base model. This delta is then used to guide the denoising process of a new base model. We evaluate DS across various SD versions, demonstrating that DS achieves consistent improvements in creating desired effects (e.g., visual styles, semantic concepts, and structures) under different sampling strategies. These results highlight DS as an effective, plug-and-play mechanism for knowledge transfer in diffusion-based image synthesis. Code:~\url{https://github.com/Zhidong-Gao/DeltaSampling}

\end{abstract}    
\section{Introduction}
\label{sec:intro}

Diffusion models~\cite{sohl2015deep,ho2020denoising,song2021denoising} have become a leading framework for image synthesis, enabling the generation of high-fidelity and semantically coherent images through iterative denoising. Large-scale text-to-image models, including DALL-E~\cite{betker2023improving}, Stable Diffusion~\cite{rombach2022high}, PixArt-$\alpha$~\cite{chen2024pixartalpha}, and Hunyuan~\cite{li2024hunyuan}, further demonstrate strong controllability and multi-modal reasoning. These capabilities enable a wide spectrum of applications, which in turn drives a demand for outputs reflecting particular artistic styles, personalized concepts, or domain-specific visual conventions, making the adaptation of pre-trained diffusion models increasingly important.

\begin{figure*}[t]
\centering
\centerline{\includegraphics[width=0.8\linewidth]{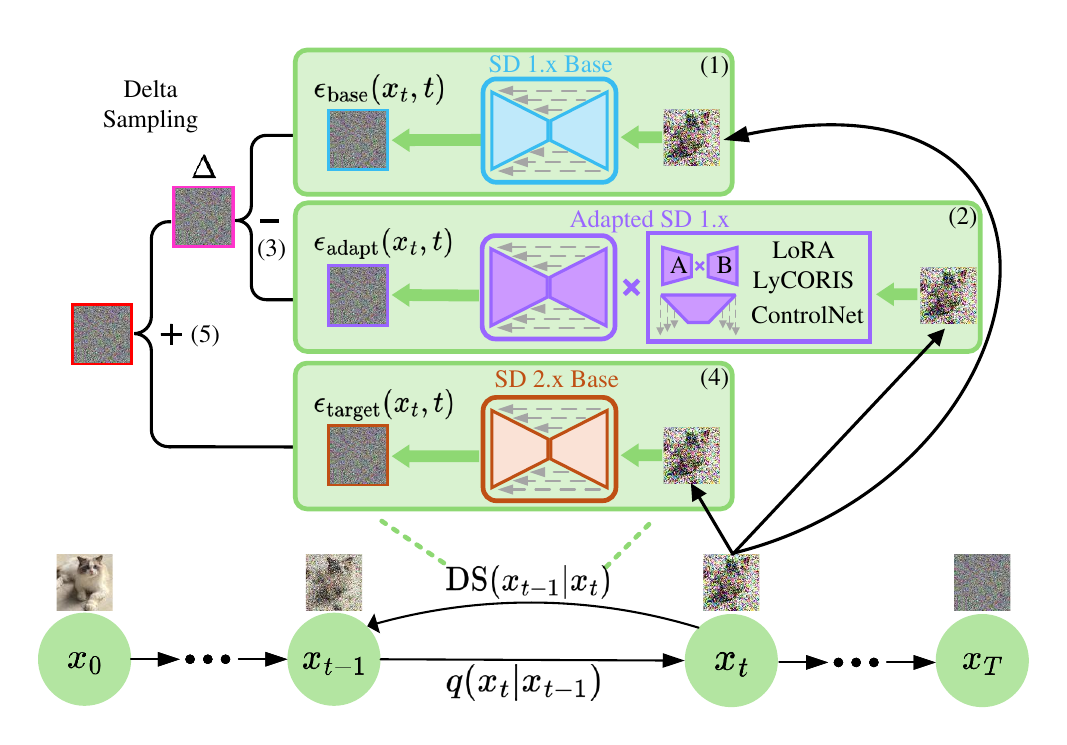}}
\caption{Overview of Delta Sampling. Each denoising step comprises the following steps: 1) The base pre-trained diffusion model predicts the noise $\epsilon_{\text{base}}(x_t,t)$ ; 2) The adapted diffusion model (Full fine-tune, LoRA, LyCORIS, ControlNet, etc.) predicts the noise $\epsilon_{\text{adapt}}(x_t,t)$; 3) The delta is computed as  $\epsilon_{\text{adapt}}(x_t,t) - \epsilon_{\text{base}}(x_t,t)$; 4) The target pre-trained diffusion model predicts the noise $\epsilon_{\text{target}}(x_t,t)$; and 5) The delta is injected into $\epsilon_{\text{target}}(x_t,t)$ to guide the denoising process of target model.}
\label{fig:framework}   
\end{figure*}

A range of adaptation methods supports such customization. One strategy is fully fine-tuning like DreamBooth~\cite{ruiz2023dreambooth}, which updates all model parameters to better accomodate new data distributions. While effective, this approach yields large checkpoint files, often several gigabytes in size, creating challenges for storage and distribution. In contrast, parameter-efficient fine-tuning methods such as LoRA~\cite{hu2022lora} and LyCORIS~\cite{yeh2023navigating} have gained popularity. These approaches inject low-rank matrices layer by layer into the frozen base model, substantially reducing trainable parameters. The resulting adapatation modules are lightweight, easy to share, and introduce no further inference latency. Another line of work involves conditioning modules such as ControlNet~\cite{zhang2023adding}, which augment the generation process by conditioning on structured inputs (e.g., edge maps, human pose) for fine-grained control over image composition. Thanks to their modular lightweight design and favorable expressiveness and controllability, platforms such as CIVITAI~\cite{CIVITAI,wei2024exploring} and HuggingFace have become the central repositories that host these open-weights modules, fostering a community-driven ecosystem that significantly reduces the barrier to customization and allows large-scale sharing of personalized diffusion models.

However, a major limitation in this ecosystem is that these adaptation modules are typically tied to the specific base model on which they were trained. When a base model evolves, such as the shift from SD 1.x to 2.x, substantial changes in network architecture, parameterization, or training distributions often occur. These shifts make previously trained adaptation modules incompatible with the new model, forcing costly retraining. Moreover, in many cases, access to the original training data is restricted or plainly unavailable, making it impractical to replicate the desired adaptations on the new base models. 

To address this challenge, we propose Delta Sampling (DS), a framework for enabling data-free knowledge transfer across diffusion models with different architectures. DS operates entirely at inference time, where it exploits the prediction difference between a base model and its adapted counterpart as a transferable signal to guide the denoising trajectory of the target model. As illustrated in Fig.~\ref{fig:framework}, at each denoising step $t$, DS first executes both a pretrained base model and its adapted counterpart (e.g., using LoRA and ControlNet) to obtain two noise predictions, denoted as $\epsilon_{\text{base}}(x_t,t)$ (Step 1) and $\epsilon_{\text{adapt}}(x_t,t)$ (Step 2). The difference between them, $\Delta := \epsilon_{\text{adapt}}(x_t,t) - \epsilon_{\text{base}}(x_t,t)$, represents the effect of adaptation (Step 3). We find that this delta encapsulates the stylistic, semantic, or structural influence from fine-tuning or conditioning, serving as a compact model-agnostic signal for that adaptation. Next, DS queries the pretrained target model to produce its own noise prediction $\epsilon_{\text{target}}(x_t,t)$ (Step 4). Finally, the delta is injected into $\epsilon_{\text{target}}(x_t,t)+\lambda \Delta$ to steer the denoising trajectory of the target model (Step 5), where $\lambda$ controls the guidance strength.

The framework is plug-and-play and agnostic to the specific form of adaptation, making it broadly applicable within the diffusion model ecosystems.  Since DS relies solely on the predicative difference between the base and adapted models, it also naturally supports integrating and transferring the combined effect of multiple adaptation components. This flexibility makes DS well-suited to real-world scenarios, where users often compose multiple modules to achieve composite generation purposes.


We evaluated the effectiveness and generality of DS across a range of adaptation methods, including fully fine-tuned checkpoints, LoRA, LyCORIS, and ControlNet. Notably, DS remains effective even when applied to IP-Adapter~\cite{ye2023ip} and Prompt Engineering~\cite{sahoo2024systematic,chen2023unleashing}, highlighting its strong model-agnostic generality. The study covers transfers across and in-between multiple SD backbones, including SD-1.5, SD-2.1, SD-XL, SD-3, and SD-3.5 (Medium and Large). In addition, we assessed DS on diverse generation tasks (artistic styles, domain-specific concepts, and structurally conditioned input) and under various sampling strategies. Across all settings, DS consistently improves the ability of target models to replicate the desired adaptation effects, demonstrating its effectiveness as a general-purpose plug-and-play solution for knowledge transfer between different diffusion models.

\section{Related Works}

\textbf{Sampling in diffusion models.}
Generating high-quality images via diffusion models typically requires dozens of iterative denoising steps~\cite{ho2020denoising}, which limits inference efficiency. DDIM~\cite{song2021denoising} introduces a deterministic non-Markovian sampler to reduce the number of steps, while PNDM and PLMS~\cite{liu2022pseudo} improve convergence through higher-order numerical integration. Distillation-based methods such as progressive distillation~\cite{salimans2022progressive} compress long sampling trajectories into a few forward passes, achieving significant speedups with minimal degradation. 

\textbf{Knowledge transfer in diffusion models.}
Transferring knowledge across domains in diffusion models remains a challenge, especially due to catastrophic forgetting during fine-tuning~\cite{hur2024expanding}. Adapter-based tuning and parameter-efficient strategies help mitigate this by preserving base model behavior~\cite{ouyang2024transfer}. Recent work further explores how knowledge is distributed across denoising steps~\cite{zhong2024diffusion}, suggesting that early-step predictions are more sensitive to adaptation. 


\textbf{Fine-tuning and adaptation.}
Numerous fine-tuning techniques exist to adapt diffusion models for new styles or subjects. DreamBooth~\cite{ruiz2023dreambooth} performs full-model tuning with prior-preservation loss for subject-driven generation. Textual Inversion~\cite{gal2023an} introduces new tokens by optimizing embedding vectors. LoRA~\cite{hu2022lora} and LyCORIS~\cite{yeh2023navigating} update low-rank matrices within the network, balancing expressiveness with parameter efficiency. ControlNet~\cite{zhang2023adding} extends the architecture with conditional branches for structured inputs such as edges, depth, or pose. Similar parameter-efficient fine-tuning~\cite{liu2024dora,LiuQFXXYF0HPWBW24,valipour-etal-2023-dylora,liu2022few,qiu2023controlling,kopiczko2024vera,buehler2024x,gao2024parameterefficient} has also been used in the diffusion model. These modular methods are widely shared via platforms like CIVITAI, but remain tied to the specific base model they were trained on. Our method enables their reuse across incompatible backbones without retraining.

\textbf{Inference-time steering in language models.}
A growing body of work in large language models explores parameter-free adaptation at decoding time. Contrastive decoding~\cite{li2023contrastive,phan2024distillation} biases generation by subtracting logits from a weaker model. DoLa~\cite{chuang2024dola} aligns output distributions by contrasting internal layers. Ensemble strategies~\cite{ormazabal2023comblm,ZhangWHQ0Z24} blend logits from small and large language models, enabling personalization and modularity. Proxy tuning~\cite{liu2024tuning} and similar methods~\cite{liu-etal-2021-dexperts,dou-etal-2019-domain,han-etal-2024-david,gera2023benefits,krause2021gedi,yang2021fudge,deng2023rewardaugmented} steer frozen models using external probability difference. These approaches demonstrate that output-level guidance at inference can effectively transfer or inject knowledge, which aligns our idea in DS framework.

\section{Preliminaries}
\subsection{Diffusion Models}
Diffusion models~\cite{sohl2015deep,ho2020denoising} are generative models that synthesize data by learning to reverse a predefined noising process. These models operate by progressively corrupting data with noise through a forward stochastic process, and then learning to reverse this process to recover the original data. Formally, given a data distribution $x_0 \sim p_{\text{data}}$, the forward diffusion process defines a Markov chain $x_0 \rightarrow x_1 \rightarrow \cdots \rightarrow x_T$, where noise is incrementally added at each step:
\begin{equation}
\begin{split}
q(x_{1:T} \mid x_0) &:= \prod_{t=1}^{T} q(x_t \mid x_{t-1}),\\
q(x_t \mid x_{t-1}) &= \mathcal{N}(x_t; \sqrt{1 - \beta_t} x_{t-1}, \beta_t \mathbf{I}),
\end{split}
\end{equation}
where $\{\beta_t\}_{t=1}^T$ is a variance schedule that determines the amount of noise added at each step. Over $T$ steps, this process gradually transforms the data into pure noise. Importantly, this formulation allows direct sampling of $x_t$ from $x_0$ with the closed-form:
\begin{equation}
q(x_t \mid x_0) = \mathcal{N}(x_t; \sqrt{\bar{\alpha}_t} x_0, (1 - \bar{\alpha}_t)\mathbf{I}),
\end{equation}
where $\alpha_t :=1-\beta_t $ and $\bar{\alpha}_t = \prod_{s=1}^t \alpha_s$.

The reverse process aims to learn a denoising model $p_\theta(x_{t-1} | x_t)$ that gradually removes noise to reconstruct $x_0$.
\begin{equation}
p_\theta(x_{0:T}) := p(x_T) \prod_{t=1}^T p_\theta(x_{t-1} \mid x_t),
\end{equation}
where \( p(x_T) = \mathcal{N}(x_T; 0, \mathbf{I}) \), and \( p_\theta(x_{t-1} \mid x_t) \) is also modeled as a Gaussian distribution.

Training is typically performed by minimizing either a variational lower bound or the following simplified objective based on noise prediction:
\[
\mathcal{L} = \mathbb{E}_{x_0, \epsilon, t} \left[ \| \epsilon - \epsilon_\theta(x_t, t) \|^2 \right],
\]
where $\epsilon \sim \mathcal{N}(0, \mathbf{I})$ and $x_t = \sqrt{\bar{\alpha}_t} x_0 + \sqrt{1 - \bar{\alpha}_t} \epsilon$. This objective encourages the model to predict the added noise $\epsilon$ from the noisy input $x_t$ and the corresponding timestep $t$.

Once the denoising model \( \epsilon_\theta \) is trained, generation proceeds by reversing the diffusion process starting from pure Gaussian noise \( x_T \sim \mathcal{N}(0, \mathbf{I}) \). This process, known as \textit{sampling}, defines a trajectory in data space and plays a central role in both the quality and speed of generation. At each timestep \( t \), the model predicts the noise component \( \epsilon_\theta(x_t, t) \), which is used to estimate the reverse conditional distribution:
\begin{equation}
x_{t-1} = \frac{1}{\sqrt{1 - \beta_t}} \left( x_t - \frac{\beta_t}{\sqrt{1 - \bar{\alpha}_t}} \epsilon_\theta(x_t, t) \right) + \sigma_t z,
\end{equation}
where $z \sim \mathcal{N}(0, \mathbf{I}),$ \( \sigma_t^2 \) is typically set to \( \beta_t \), matching the forward process variance. 

While the above sampling procedure is effective, it is computationally expensive due to the large number of diffusion steps. Subsequent work has introduced more efficient sampling methods, including deterministic approximations such as DDIM~\cite{song2021denoising}, and higher-order numerical solvers~\cite{karras2022elucidating}, such as Euler, Heun~\cite{karras2022elucidating}, and DPM-Solver~\cite{lu2022dpm}. These methods significantly reduce inference time while maintaining high sample quality, making them widely adopted in diffusion ecosystems.

\subsection{Adaptation Techniques}
\paragraph{LoRA: Low-Rank Adaptation.}
LoRA~\cite{hu2022lora} is a parameter-efficient fine-tuning technique that adapts large pre-trained models by injecting trainable low-rank matrices into selected layers while keeping the original weights frozen. Specifically, given a weight matrix \( W \in \mathbb{R}^{m \times n} \), LoRA introduces the trainable parameters in the form:
\begin{equation}
W' = W + \Delta W, \quad \text{where} \quad \Delta W = B A,
\end{equation}
where \( A \in \mathbb{R}^{r \times n} \) and \( B \in \mathbb{R}^{m \times r} \) are two low-rank matrices (\( r \ll \min(m, n) \)). Only \( A \) and \( B \) are updated during fine-tuning. This design yields compact, shareable adaptation modules that impose negligible overhead at inference time and are widely used for diffusion models customization.
\paragraph{LyCORIS: Other Low-Rank Adaptations.} LyCORIS~\cite{yeh2023navigating} represents the Lora beYond Conventional methods, Other Rank adaptation Implementations for Stable diffusion. It extends the conventional  LoRA paradigm by introducing more expressive and flexible matrix structures for parameter-efficient fine-tuning. One notable variant within LyCORIS is LoHa (Low-Rank Adaptation with Hadamard Product), which enhances the expressiveness of LoRA by replacing the single low-rank matrix product with a composition of two low-rank products, combined via the Hadamard product. Formally, given a frozen weight matrix \(W\), LoHa introduces trainable parameters in the following form:
\begin{equation}
W' = W + \Delta W, \quad \text{where} \quad \Delta W = (B_1 A_1) \odot (B_2 A_2),
\end{equation}
where \( A_1, A_2 \in \mathbb{R}^{r \times n} \) and \( B_1, B_2 \in \mathbb{R}^{m \times r} \) are trainable low-rank matrices, and \( \odot \) denotes the Hadamard product. In standard LoRA, the resulting matrix of $BA$ has rank at most r. In contrast, LoHa can achieve a maximum rank of $r^2$, significantly expanding the expressive power of the adaptation  module, thereby increasing the expressive capacity given the same number of trainable parameters. As with LoRA, LyCORIS is widely adopted for efficient and flexible fine-tuning of diffusion models.
\paragraph{ControlNet: Structure-Conditioned Adaptation.}
ControlNet~\cite{zhang2023adding} extends SD model by enabling conditioning on external structural inputs, such as edge maps, depth maps, or human poses. It achieves this by duplicating the encoder of the original U-Net and inserting additional convolutional layers, which are initialized to zero. These layers are trained to process the conditioning input $c$ and guide the generation process. Meanwhile, the original U-Net remains frozen, preserving the model's original capabilities while allowing condition-aware generation.

Let \( x_t \) be the noisy latent at timestep \( t \), and let \( \phi_{\text{base}}^l \) denote the activation of the frozen base U-Net's encoder at layer \( l \). The control branch computes \( \phi_{\text{ctrl}}^l(c^{\prime}) \), which is injected into input of the U-Net's decoder,
\begin{equation}
\tilde{\phi}^l = \phi_{\text{base}}^l + \phi_{\text{ctrl}}^l(c^{\prime}).
\end{equation}
Here $c^{\prime}$ denotes the image conditioning for ControlNet. This mechanism enables precise, spatially-aware conditioning without modifying the original model, supporting flexible and modular generation pipelines.

\paragraph{Challenges in Cross-Version Reuse.} Although the adaptation techniques discussed above provide flexible and parameter-efficient means for fine-tuning diffusion models, they are fundamentally constrained by the architectural characteristics of the base models on which they are trained. This architectural dependency substantially limits the reusability of adaptation modules across different versions of diffusion models.

Different model versions are commonly trained using distinct training pipelines, including variations in optimization strategies, data distributions, and network configurations. Consequently, their parameter spaces differ in both dimensionality and functional representation. For example, the cross-attention layers in SD-1.5 and SD-2.1 have dimensions of 768 and 1024, respectively. As a result, LoRA modules optimized for SD-1.5 cannot be directly applied to SD-2.1 due to mismatched weight shapes. LyCORIS modules face analogous limitations, as their low-rank factorization is inherently tied to the original parameter dimensions. Fully fine-tuned checkpoints also cannot be transferred across model versions because architectural discrepancies prevent direct parameter alignment. Similarly, ControlNet depends on version-specific intermediate feature map resolutions and projection matrices, making their direct reuse infeasible.

The challenge becomes more pronounced when multiple adaptation modules are used jointly. In practice, users often stack several LoRA, LyCORIS, ControlNet, and other auxiliary components on top of a fully fine-tuned model to achieve complex generative behaviors. These components are typically trained on heterogeneous datasets, under distinct objectives and hyperparameter configurations, making unified transfer to a new base model highly non-trivial. A potential strategy for cross-version reuse is knowledge distillation, where a new model is trained to replicate the composite behavior of the original adapted system. However, this process is computationally expensive, requires careful system design, and often depends on access to the original training data, which may be unavailable in many real-world scenarios. Consequently, in practice, cross-version reuse of adaptation modules remains difficult and frequently infeasible.

\section{Our Method}
To enable knowledge transfer across diffusion models with different architectures, we propose Delta Sampling, a plug-and-play framework that injects guidance signals from an adapted model into the denoising trajectory of a different target model. Our key insight is that the behavioral difference between a base model and its adapted variant implicitly encodes the effect of fine-tuning or conditioning, which can be reused as a transferable signal. By extracting this difference during the sampling process and injecting it into the target model, we can steer generation toward desired styles, semantics, or control objectives---\textit{without any training or architecture constraints}.

Let \( \theta \) denote the parameters of a pre-trained diffusion model (e.g., Stable Diffusion 1.5). Fine-tuning this model using different adaptation strategies results in new parameterizations:
\begin{itemize}
    \item \( \theta' \): parameters after full-model fine-tuning
    \item \( \theta_{\text{LoRA}} \): LoRA weights
    \item \( \theta_{\text{LyCORIS}} \): extended low-rank adaptation (e.g., LoHa)
    \item \( \theta_{\text{ControlNet}} \): auxiliary parameters for ControlNet
\end{itemize}

Each of these represents a distinct form of adaptation. In practice, these methods are often composed: for example, a LoRA adapter and a ControlNet module may be used simultaneously. We define the adapted model \( \mathcal{A} \) as:
\begin{equation}
\begin{split}
&\mathcal{A} := 
\begin{cases}
\theta' \cup \Theta_{\text{aux}}, & \text{if full fine-tuning}, \\
\theta \cup \Theta_{\text{aux}}, & \text{otherwise},
\end{cases} \\
&\text{where} \quad \Theta_{\text{aux}} \subseteq \{ \theta_{\text{LoRA}},\ \theta_{\text{LyCORIS}},\ \theta_{\text{ControlNet}}\}.
\end{split}
\end{equation}
In our formulation, \( \mathcal{A} \) refers to an adapted model composed of a base model or its full-fined variant \( \{ \theta , \theta' \}\) and an optional set of auxiliary modules \( \Theta_{\text{aux}} \subseteq \{\theta_{\text{LoRA}},\ \theta_{\text{LyCORIS}},\ \theta_{\text{ControlNet}}\} \). This definition allows for flexible adaptation configurations, including full fine-tuning checkpoints, LoRA and LyCORIS adapters, and ControlNet, or combinations of multiple components. Notably, in the special case where \( \mathcal{A} = \theta \), i.e., the adapted model is identical to the base model, the guidance signal vanishes at all timesteps, reducing DS to the standard sampling procedure of the target model.

Let \( \epsilon_{\mathcal{A}}(x_t, t \mid c^{\prime}) \) be the noise prediction at timestep \( t \) from the adapted model, conditioned on \( c^{\prime} \) (e.g., human pose and text prompt). Likewise, let \( \epsilon_{\theta}(x_t, t \mid c) \) denote the prediction of the original base model without adaptation, conditioned on \( c \) (e.g., text prompt only).

\paragraph{Residual Extraction and Interpretation.}
We define the \emph{residual signal} between the adapted model and its corresponding base model as:
\begin{equation}
\Delta(x_t, t) := \epsilon_{\mathcal{A}}(x_t, t \mid c^{\prime}) - \epsilon_{\theta}(x_t, t \mid c).
\end{equation}

This residual represents a \emph{behavioral delta}: a direct measure of how the adaptation $\mathcal{A}$ modifies the denoising dynamics of the base diffusion model at each timestep. Rather than relying on architectural or parameter-level differences, $\Delta$ captures the \emph{functional effect} of the adaptation—encoding stylistic, semantic, or structural influences imparted through fine-tuning or auxiliary conditioning. In this sense, it approximates the first-order change in the model’s generative trajectory induced by the adaptation.

A key advantage of this formulation is that the residual operates entirely in the prediction space of the model, making it inherently portable. Unlike weight parameters, which are tightly bound to a specific architecture or training configuration, the behavioral delta can be injected into a structurally different target model without requiring retraining, fine-tuning, or weight alignment. This enables DS to transfer adaptation effects across heterogeneous diffusion backbones purely at inference time, providing a compact and model-agnostic mechanism for cross-model knowledge transfer.

\paragraph{Guided Sampling with Delta.}

\begin{algorithm}[t]
\caption{Delta Sampling}
\label{alg:delta_sampling}
\KwIn{Target model \( \mathcal{T} \), base model \( \theta \), adapted model \( \mathcal{A} \), conditioning \( c \) and \( c^{\prime} \), number of steps \( T \), variance schedule $\{\beta_t\}_{t=1}^T$, guidance strength \( \lambda \)}
\KwOut{Generated sample \( x_0 \)}

Sample \( x_T \sim \mathcal{N}(0, \mathbf{I}) \) \tcp*[f]{Initialize with pure Gaussian noise}

\For{\( t = T \) \KwTo \( 1 \)}{
    \( \epsilon_{\text{base}} \gets \epsilon_{\theta}(x_t, t \mid c) \) \tcp*[f]{Predict noise using the base model}
    
    \( \epsilon_{\text{adapt}} \gets \epsilon_{\mathcal{A}}(x_t, t \mid c^{\prime}) \) \tcp*[f]{Predict noise using the adapted model}
    
    \( \Delta \gets \epsilon_{\text{adapt}} - \epsilon_{\text{base}} \) \tcp*[f]{Compute the delta}
    
    \( \epsilon_{\text{target}} \gets \epsilon_{\mathcal{T}}(x_t, t \mid c) \) \tcp*[f]{Get the target model's noise prediction}
    
    \( \hat{\epsilon} \gets \epsilon_{\text{target}} + \lambda \cdot \Delta \) \tcp*[f]{Inject the delta as a guiding signal}
    
    Update \( x_{t-1} \) using equation~(\ref{equ:new_ddpm}) with \( \hat{\epsilon} \) \tcp*[f]{Perform denoising step}
}
\Return \( x_0 \) \tcp*[f]{Final generated sample}
\end{algorithm}
Given a target model \( \mathcal{T} \), our goal is to guide its denoising trajectory using \( \Delta(x_t, t) \), without modifying its parameters. Let \( \epsilon_{ \mathcal{T}}(x_t, t \mid c) \) be the target model’s prediction. We define the \textit{Delta-guided noise prediction} as:
\begin{equation}
\hat{\epsilon}(x_t, t) = \epsilon_{\mathcal{T}}(x_t, t \mid c) + \lambda \cdot \Delta(x_t, t),
\end{equation}
where \( \lambda \in [0, \infty) \) is a user-defined scalar that controls the strength of guidance.

We then plug \( \hat{\epsilon}(x_t, t) \) into the standard sampling process:
\begin{equation}
x_{t-1} = \frac{1}{\sqrt{1 - \beta_t}} \left( x_t - \frac{\beta_t}{\sqrt{1 - \bar{\alpha}_t}} \hat{\epsilon}(x_t, t) \right) + \sigma_t z.
\end{equation}\label{equ:new_ddpm}
This procedure is repeated for each \( t = T, T-1, \dots, 1 \). An overview is summarized in Algorithm~\ref{alg:delta_sampling}. While the formulation (\ref{equ:new_ddpm}) is based on the DDPM ancestral sampler, the proposed framework is fully compatible with a wide range of alternative sampling strategies. Since our method operates directly on the predicted noise estimates \( \epsilon(x_t, t) \), it can be seamlessly integrated into other samplers such as DDIM~\cite{song2021denoising}, Euler, or Heun~\cite{karras2022elucidating} without requiring any modification to their underlying update rules. In all cases, the predicted noise from the target model is simply augmented by the residual signal extracted from the base-adapted pair. This generality makes DS a flexible and modular inference-time technique for knowledge transfer, agnostic to the specific choice of sampling algorithm.

\paragraph{Guidance Strength.}
The scalar hyperparameter $\lambda$ governs the influence of the transferred adaptation on the target model. It acts as a multiplicative factor applied to the residual signal $\Delta(x_t, t)$ before this signal is added to the target model’s noise prediction. When $\lambda = 0$, DS becomes inactive, and the target model $\mathcal{T}$ samples purely from its own learned distribution. As $\lambda$ increases, the behavioral contribution of the adapted model $\mathcal{A}$—as encoded in the residual—becomes progressively more prominent, enabling a controllable interpolation between the native generative tendencies of $\mathcal{T}$ and the stylistic, semantic, or structural attributes captured by the adaptation.

The optimal choice of $\lambda$ can vary across tasks, depending on the strength and specificity of the underlying adaptation. Empirically, we find that moderate values (typically $\lambda \in [0.5, 1.5]$) strike a good balance between faithfully transferring the desired effects and preserving the visual fidelity of the target model. Excessively large values may over-amplify the residual and destabilize generation, while very small values may result in insufficient adaptation transfer. Thus, $\lambda$ provides a simple yet effective mechanism to adjust the degree of influence exerted by the source adaptation during sampling.

\paragraph{$\lambda$-Decay.}
While DS typically uses a fixed guidance strength $\lambda$, we find that applying a constant value across all denoising steps may lead to instability, particularly when the discrepancy between the adapted model and the target model is large. In practice, strong guidance is most beneficial during early denoising stages, whereas excessive late-stage guidance may introduce artifacts or disrupt fine-detail refinement. To address this, we adopt a timestep-dependent \emph{$\lambda$-decay} strategy that gradually reduces the guidance strength as sampling progresses. This improves stability and preserves image quality without altering the underlying DS formulation. We refer readers to Appendix~\ref{appendix:lambda-decay} for the full details.

\paragraph{Multi-Module Adaptation and Composition.}
A notable advantage of Delta Sampling is its ability to seamlessly transfer the \emph{combined influence of multiple adaptation modules} within a single unified framework. In practical workflows, users frequently compose several adaptation strategies---such as LoRA or LyCORIS for stylistic refinement, ControlNet for structural conditioning, and even fully fine-tuned checkpoints for domain specialization---to achieve complex generative objectives. Importantly, DS requires no modification or disentangling of these components.

Because the residual signal \( \Delta(x_t, t) \) is computed directly from the prediction discrepancy between the adapted model and its corresponding base model, it inherently captures the \emph{aggregate effect} of all applied modules. Regardless of whether these effects arise from low-rank adaptations, auxiliary conditioning branches, or large-scale parameter shifts, their combined behavioral influence is encoded in the residual and transferred as a unified guiding signal. This property enables DS to naturally support multi-module compositions without explicit architectural alignment or module-specific integration logic.

In practice, this capability greatly enhances the flexibility and expressiveness of model transfer. Community-created checkpoints often stack several adaptation components to achieve highly customized generation behaviors, and DS allows these rich compositions to be reused directly on different target backbones---\emph{without retraining, merging weights, or accessing the original data}. This plug-and-play compatibility with arbitrarily complex adaptation stacks underscores the generality and practicality of DS as a universal inference-time knowledge transfer mechanism.

\begin{figure*}[ht]
\centering
\centerline{\includegraphics[width=0.95\linewidth]{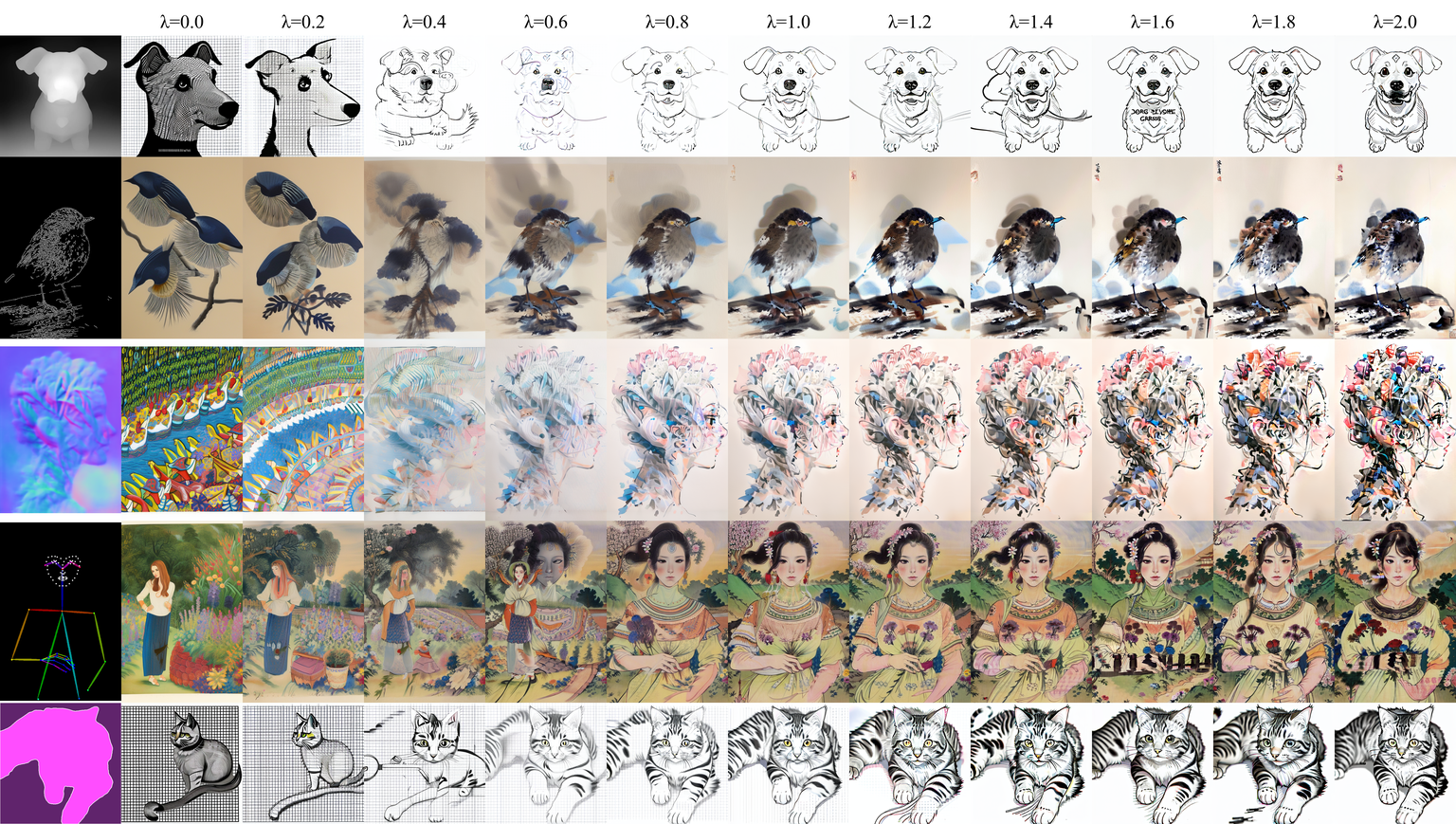}}
\caption{DS with full-fined checkpoint, LoRA and ControlNet. From top to bottom, the results correspond to different control conditions: depth (first rows), canny edge (second rows), human pose (second rows), and segmentation (bottom rows).}
\label{fig:controlnet_transfer}
\end{figure*}

\section{Experiment}


To comprehensively assess the effectiveness, robustness, and generality of Delta Sampling (DS), we conduct extensive experiments across a wide spectrum of adaptation modules, generation tasks, and diffusion backbones. Our evaluation spans multiple pairs of base and target models. Specifically, we examine the following representative backbones:
\textbf{SD-1.5}\footnote{\url{https://huggingface.co/stable-diffusion-v1-5/stable-diffusion-v1-5}};
\textbf{SD-2.1}\footnote{\url{https://stability.ai/news/stablediffusion2-1-release7-dec-2022}};
\textbf{SD-3}\footnote{\url{https://huggingface.co/stabilityai/stable-diffusion-3-medium}};
\textbf{SD-3.5 Medium}\footnote{\url{https://huggingface.co/stabilityai/stable-diffusion-3.5-medium}};
\textbf{SD-3.5 Large}\footnote{\url{https://huggingface.co/stabilityai/stable-diffusion-3.5-large}}.

For adaptation modules, we utilize a diverse collection of publicly available community-generated assets from CIVITAI, including fully fine-tuned checkpoints, LoRA and LyCORIS adapters, and ControlNet models. All modules were originally trained on their respective source backbones and are therefore not directly compatible with the target models due to architectural differences, parameterization changes, and mismatched training distributions. DS enables these modules to be reused without retraining, providing a direct measure of its zero-shot transfer capability. These settings cover transfers across distinct architectural generations, latent spaces, and training paradigms, allowing us to evaluate DS under both moderate and substantial model shifts.

Unless otherwise specified, we adopt a consistent experimental protocol across all model pairs. Images are generated at a resolution of $512 \times 512$ or $512 \times 768$, using 16 denoising steps with the Euler sampler and a standard variance schedule. The random seed is fixed to 42 for reproducibility. For experiments involving LoRA or LyCORIS, the adapter scaling factor is set to 1.0. To ensure stable generation quality across all evaluations, we apply the following negative prompt:
\texttt{``worst quality, low quality, symbol, text, logo''}. For reproducibility, the complete sampling workflows used to generate the figures in the experiment, including LoRA loading configurations, ControlNet preprocessing, and DS-specific sampling scripts, are provided in our repository\footnote{\url{https://github.com/Zhidong-Gao/DeltaSampling}}.

\paragraph{Quantitative Evaluation on LoRA and LoHa Adaptations.}
To quantitatively assess the effectiveness of DS in transferring parameter-efficient adaptations from SD-1.5 to SD-2.1, we evaluate both \textit{similarity} and \textit{diversity} using the standard LoRA/LoHa benchmark introduced in prior work~\cite{yeh2023navigating}. We compare four settings: (1) SD-2.1 without any adaptation (\textit{SD-2.1 Only}), (2) DS applied to SD-2.1 using a 32-rank, 16-alpha LoRA adapter (\textit{DS w/ LoRA}), (3) DS applied to SD-2.1 using a 16-rank, 8-alpha LoHa adapter (\textit{DS w/ LoHa}), and (4) SD-1.5 with the original LoRA adapter (\textit{SD-1.5 + LoRA}), which serves as the upper bound representing the desired target effect.

Following the evaluation protocol in~\cite{yeh2023navigating}, \textbf{Similarity} is measured using CLIP-based metrics that quantify adherence to the target style or concept, comparing generated images against reference prompts or ground-truth exemplars. \textbf{Diversity} is assessed using the average LPIPS distance computed between image pairs generated from the same prompt. Higher similarity indicates stronger stylistic fidelity, while higher diversity reflects richer visual variation without mode collapse.

\begin{figure*}[ht]
    \centering
    \begin{subfigure}[t]{0.45\linewidth}
        \centering
        \includegraphics[width=\linewidth]{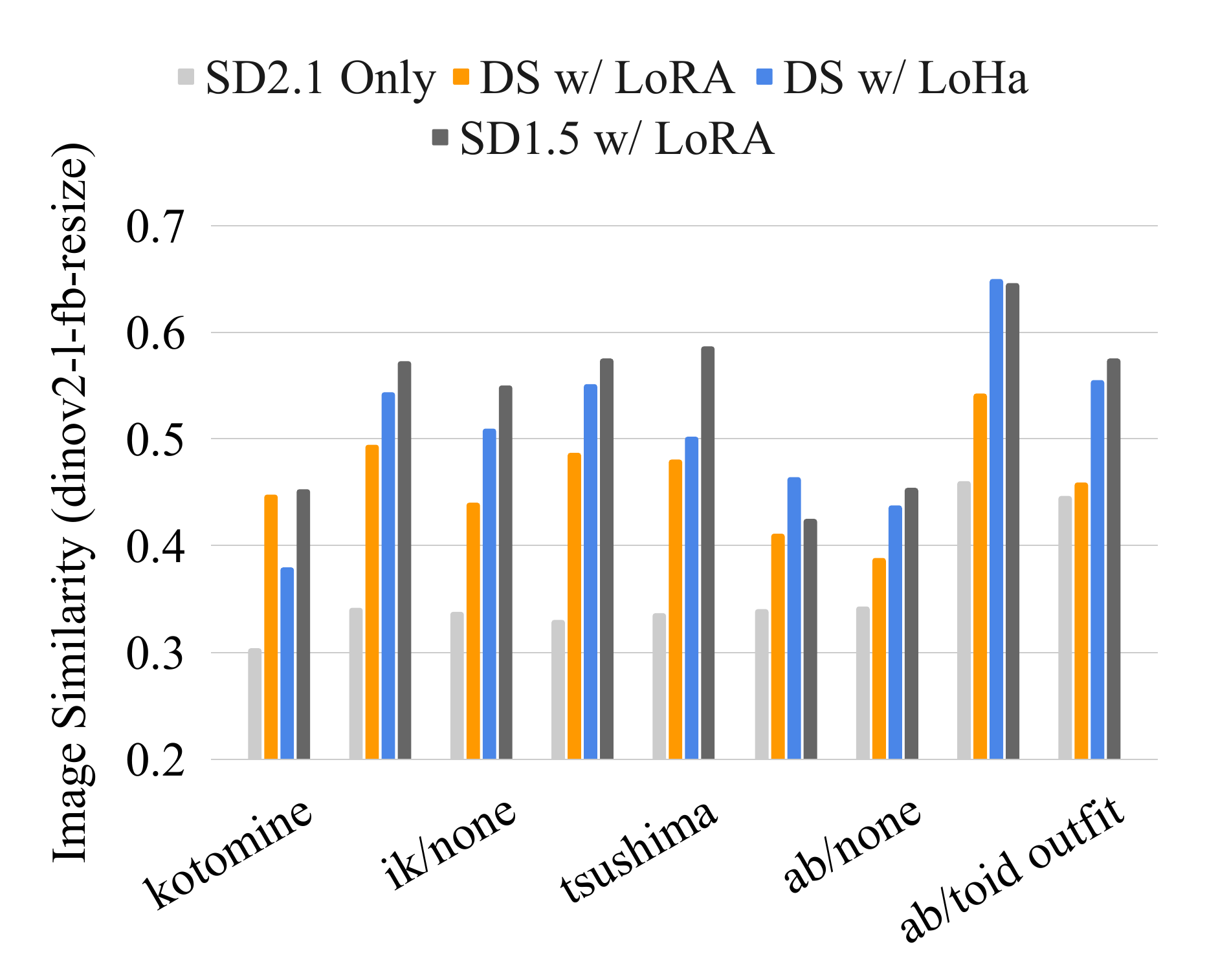}
        \caption{Similarity of category "Anime"}
        \label{fig:sub1}
    \end{subfigure}
    \begin{subfigure}[t]{0.45\linewidth}
        \centering
        \includegraphics[width=\linewidth]{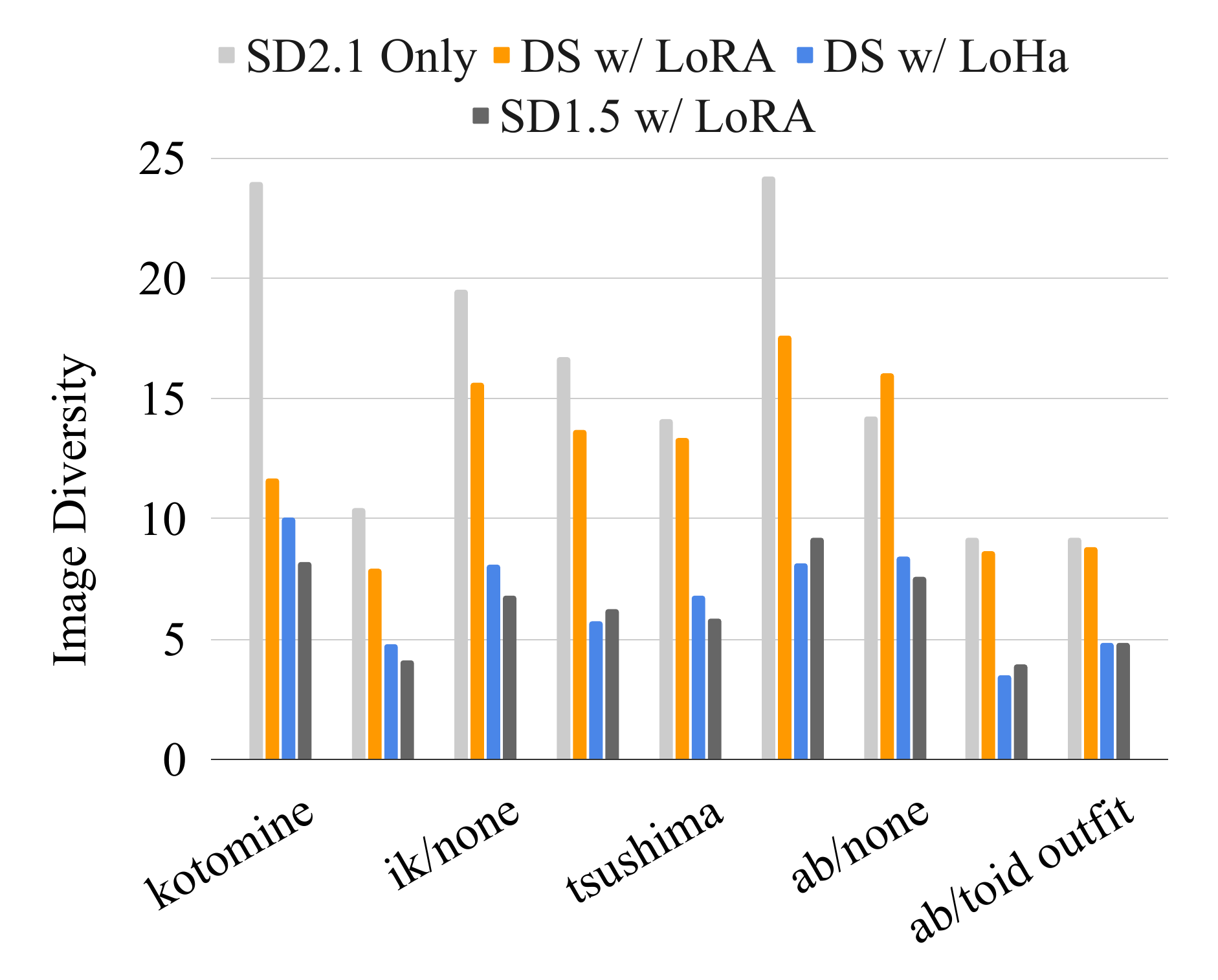}
        \caption{Diversity of category "Anime"}
        \label{fig:sub2}
    \end{subfigure}
    \caption{We compare: (1) SD-2.1 Only (baseline, no adaptation), (2) DS applying a 32-rank, 16-alpha LoRA (DS w/ LoRA) to SD-2.1, (3) DS applying a 16-rank, 8-alpha LoHa (DS w/ LoHa) to SD-2.1, and (4) SD-1.5 with the original 32-rank, 16-alpha LoRA (baseline, representing the desired target effect). Similarity measures adherence to the target style/concept (e.g., using CLIP-Score with reference prompts/images), while diversity assesses visual variation among images generated for the same prompt (e.g., using average LPIPS between pairs).}
    \label{fig:combined}
\end{figure*}

As shown in Fig.~\ref{fig:combined}, DS substantially improves similarity scores over the SD-2.1 baseline for both LoRA and LoHa modules, confirming that the stylistic and conceptual characteristics encoded in SD-1.5 adapters are successfully transferred. At the same time, DS maintains LPIPS diversity, demonstrating that guiding SD-2.1 with residual signals preserves output variability. These results highlight the ability of DS to faithfully and robustly transfer a wide range of lightweight adaptation modules to architecturally different target backbones.

In addition to the main quantitative results, the Appendix~\ref{sec:exp_more_lyco} includes an expanded set of evaluations on LoRA and LoHa adaptations. Specifically, we report per-category similarity and diversity measurements under different guidance strengths~(\(\lambda\)), covering the \textit{scene}, \textit{anime}, \textit{people}, \textit{style}, and \textit{toy} categories. These supplementary results provide a more complete view of DS performance across diverse semantic domains and guidance settings.

\paragraph{Control Transfer via Multi-Module Compositions with ControlNet.}
In Fig.~\ref{fig:controlnet_transfer}, we evaluate DS under a significantly more challenging multi-module composition setting that involves the \emph{simultaneous} transfer of full fine-tuned checkpoints, LoRA modules, and ControlNet-based structural conditioning. We experiment with five widely used ControlNet modalities: \textit{depth}, \textit{Canny edge}, \textit{surface normal}, \textit{human pose}, and \textit{semantic segmentation}. While these ControlNet variants typically share a similar architectural backbone, they differ substantially in their input modalities and the type of structural information they encode, and all are originally trained on SD-1.5. Direct weight reuse on SD-2.1 is therefore not possible due to parameter and feature-space mismatches.

Despite the heterogeneity of the conditioning signals and their tight coupling to the SD-1.5 feature space, DS enables SD-2.1 to reproduce their combined effects in a coherent manner. The generated images faithfully retain: \emph{(i)} the global stylistic priors introduced by the full fine-tuned checkpoint, \emph{(ii)} the fine-grained appearance traits contributed by LoRA, and \emph{(iii)} the structural constraints enforced by the ControlNet module. Importantly, DS achieves this without explicit disentanglement or module-specific adaptation---the residual signal implicitly aggregates all behavioral modifications arising from the stacked modules.

These results demonstrate that DS is capable of transferring complex, multi-component adaptation stacks that jointly encode style, semantics, and geometry. This highlights the method’s strong compositionality and its practical applicability to real-world community workflows, where multiple adaptation modules are often combined to achieve rich and highly customized generation behavior.

\paragraph{Additional Results in the Appendix.}
We include a broad set of supplementary experiments in the Appendix, further demonstrating the versatility and generality of Delta Sampling (DS). Specifically, the Appendix provides:

\begin{itemize}
    \item \textbf{(\ref{sec:exp_full_fine_tune})} Additional results on transferring full fine-tuned checkpoints, showing that DS successfully preserves global stylistic priors across architectures.

    \item \textbf{(\ref{sec:exp_sample})} An expanded evaluation of DS under diverse diffusion samplers, confirming that the residual-guidance mechanism remains stable across both deterministic and stochastic solvers.

    \item \textbf{(\ref{sec:exp_15_xl})} Bidirectional transfer of ControlNet-based structural conditioning between \textbf{SD-1.5} and \textbf{SD-XL}, illustrating DS’s ability to transplant structural priors across substantially different backbones.

    \item \textbf{(\ref{sec:exp_3_35})} Cross-generation ControlNet transfer among \textbf{SD-3}, \textbf{SD-3.5 Medium}, and \textbf{SD-3.5 Large}, demonstrating compatibility with modern diffusion families.

    \item \textbf{(\ref{sec:exp_ip})} Transfer of IP-Adapter conditioning from \textbf{SD-1.5} to \textbf{SD-2.1}, highlighting DS’s applicability to image-based conditioning.

    \item \textbf{(\ref{sec:exp_prompt})} Transfer of prompt-engineered behaviors from \textbf{SD-3.5-Medium} to \textbf{SD-3.5-Large}, even when target prompts lack corresponding stylistic tokens.


    \item \textbf{(\ref{sec:exp_mult_adapt})} Demonstrations of DS’s compositional capabilities, including:  
    \textit{(i)} joint transfer of \textbf{multiple ControlNet modules} within a single generation;  
    \textit{(ii)} composition of \textbf{multiple LoRA adapters} to achieve blended stylistic or semantic effects; and  
    \textit{(iii)} cross-model LoRA composition, where LoRAs originating from different base models are simultaneously transferred to a shared target architecture. These results showcase DS’s ability to support complex, real-world adaptation pipelines.
    
\end{itemize}

\section{Conclusion}
Based on the analysis of the adaptation process and its effect on denoising step predictions, we have introduced a simple and effective knowledge transfer method, Delta Sampling (DS). The proposed delta captures the stylistic, semantic, or structural influence of an adaptation, serving as a model-agnostic guidance signal. To transfer this knowledge, we develop an effective inference-time algorithm that injects this delta directly into the denoising process of a new target model. The proposed algorithm does not require costly retraining steps, access to the original training data, or an update to model weights. While our evaluation demonstrates broad applicability across diverse models and tasks, future work could include deeper quantitative analysis to optimize, understand, and extend the DS mechanism.


{
    \small
    \bibliographystyle{ieeenat_fullname}
    \bibliography{main}
}

\clearpage
\appendix
\twocolumn[
\begin{center}
\Large\bfseries Appendix
\end{center}
]
\section{Timestep-Dependent Guidance Scheduling}
\label{appendix:lambda-decay}
\paragraph{$\lambda$-Decay.}
Although a fixed guidance strength $\lambda$ provides a straightforward mechanism to regulate the influence of the adapted model, we observe that maintaining a constant value throughout the entire denoising trajectory may cause instability, especially when there exists a substantial discrepancy between the adapted model and the target model. During early denoising stages, when $x_t$ remains close to pure noise, a strong residual signal is generally beneficial as it effectively shapes the global structure and high-level semantics required for reproducing the desired adaptation. However, in later stages, the denoising process becomes increasingly sensitive to perturbations. Excessive guidance in this phase may interfere with the target model’s intrinsic priors, leading to artifacts such as exaggerated high-frequency details, irregular texture amplification, or degraded local coherence.

To enhance stability while preserving the adaptation effect, we introduce a timestep-dependent guidance schedule $\lambda(s)$, where $s \in [0,1]$ is a normalized time variable derived from the discrete denoising timestep $t \in \{1,\ldots,T\}$. Specifically, we define:
\begin{equation}
    s = \frac{t-1}{T-1},
\end{equation}
which monotonically maps the denoising trajectory from coarse-to-fine steps onto a continuous interval, with $s = 0$ corresponding to the earliest denoising stage and $s = 1$ corresponding to the final refinement stage. The noise estimate is then modified as:
\begin{equation}
    \hat{\epsilon}(x_t, t)
    = \epsilon_{\mathcal{T}}(x_t, t \mid c)
    + \lambda(s) \cdot \Delta(x_t, t),
\end{equation}
allowing the residual signal to guide global structural formation while ensuring that high-frequency details are refined primarily by the target model.

We instantiate $\lambda(s)$ using several lightweight and practical decay functions:
\begin{itemize}
    \item \textbf{Linear decay:}
    \begin{equation}
        \lambda(s) = \lambda_{\max}
        - (\lambda_{\max} - \lambda_{\min}) \cdot s.
    \end{equation}

    \item \textbf{Exponential decay:}
    \begin{equation}
        \lambda(s) =
        \lambda_{\min}
        + (\lambda_{\max} - \lambda_{\min}) e^{-k s},
    \end{equation}
    which preserves strong early-stage guidance while rapidly suppressing the residual signal near the final steps.

    \item \textbf{Cosine decay:}
    \begin{equation}
        \lambda(s) =
        \lambda_{\min}
        + \frac{1}{2}(\lambda_{\max} - \lambda_{\min})(1 + \cos(\pi s)),
    \end{equation}
    providing a smooth and symmetric transition that avoids abrupt changes in guidance strength.
\end{itemize}

These schedules are sampler-agnostic and incur negligible computational overhead. In our experiments, cosine decay offers particularly stable behavior due to its smooth attenuation profile, whereas exponential decay is advantageous for adaptations that introduce strong or localized residual signals (e.g., ControlNet or highly stylized LoRA configurations).

\section{Transfer Result of Full Fine-Tuned Checkpoints} \label{sec:exp_full_fine_tune}
\begin{figure*}[ht]
\centering
\centerline{\includegraphics[width=\linewidth]{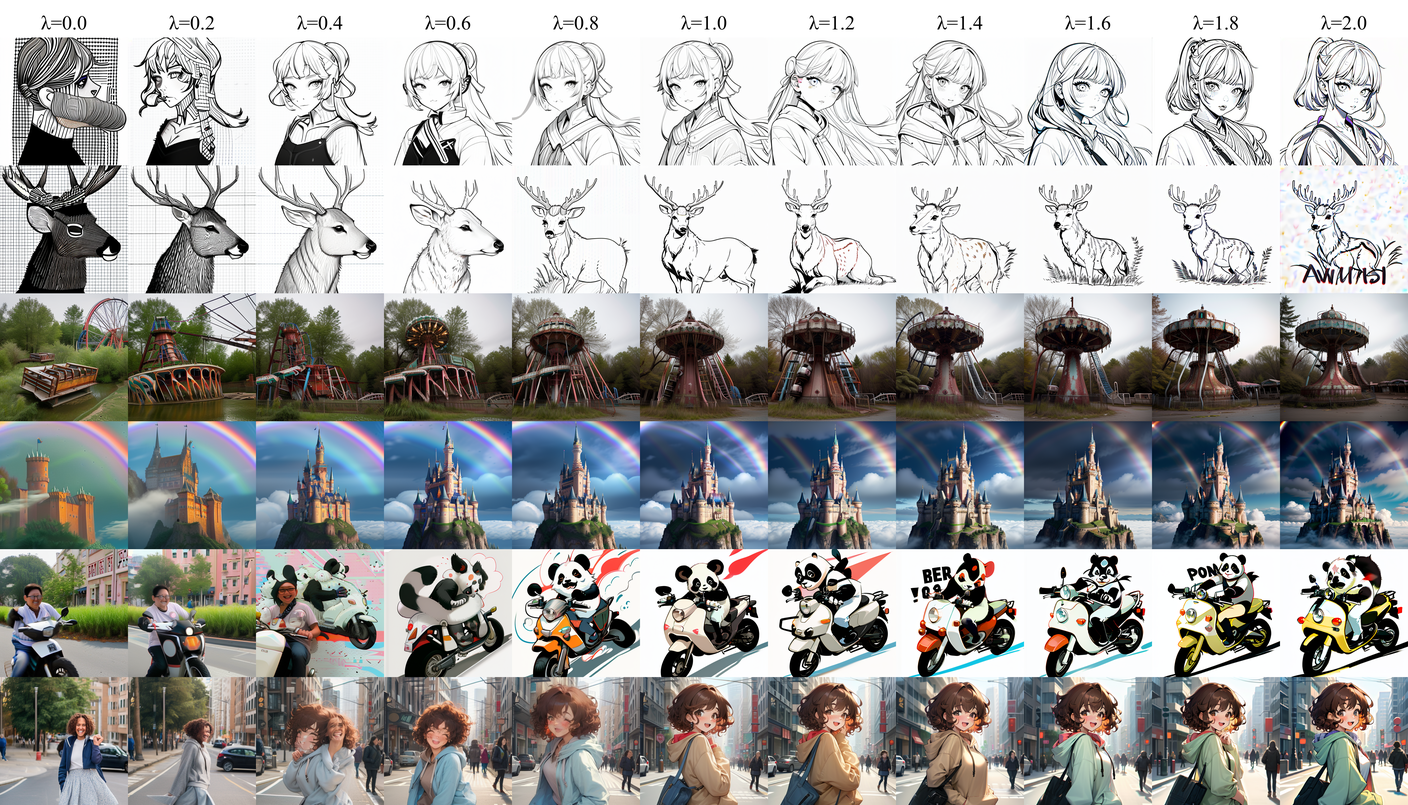}}
\caption{DS with full-fined checkpoint. From top to bottom, the results correspond to three different checkpoints: \texttt{LineArt.safetensors} (first two rows), \texttt{Photon\_v1.safetensors} (middle two rows), and \texttt{revAnimated\_v2.safetensors} (bottom two rows). From left to right, each column shows the result generated with a different guidance strength $\lambda$.}
\label{fig:full_transfer}
\end{figure*}
We first evaluate DS in the setting of stylistic and conceptual transfer using full-model fine-tuned checkpoints and LoRA-based adaptations originally trained on SD-1.5. Representative qualitative results for the SD-1.5~$\rightarrow$~SD-2.1 transfer are shown in Fig.~\ref{fig:full_transfer}. Each row corresponds to a distinct adapted model, and each column displays outputs generated with a different guidance strength $\lambda \in \{0.0, 0.2, \dots, 2.0\}$. The leftmost column ($\lambda = 0$) serves as the baseline, showing unguided SD-2.1 outputs without any transferred adaptation. The first two rows illustrate the transfer of an anime-style line-art checkpoint (\texttt{LineArt.safetensors}) under prompts such as \texttt{1girl, lineart} and \texttt{deer, lineart}. The next two rows present photorealistic fine-tuned models with prompts like \texttt{abandoned park} and \texttt{castle in cloud}, while the final rows show results from anime-style LoRA modules. Across all cases, SD-2.1 alone is unable to reproduce the corresponding visual style or object characteristics (i.e., $\lambda = 0$). As the guidance strength increases, the target model gradually incorporates the stylistic and semantic cues encoded in the residual signal, yielding outputs that more closely match the intended appearance. Effective transfer typically emerges for $\lambda \geq 0.6$, while overly large values (e.g., $\lambda > 1.6$) may introduce visual distortions or oversaturated stylistic artifacts. These results demonstrate that DS provides a robust and controllable mechanism for reusing full fine-tuned checkpoints and LoRA-based adaptations across architecturally different diffusion models. Even when the original weights are incompatible with the target backbone, DS preserves stylistic identity and concept structure through purely inference-time residual guidance.

\section{Transfer Result under Different Samplers}\label{sec:exp_sample}
\begin{figure*}[ht]
\centering
\centerline{\includegraphics[width=\linewidth]{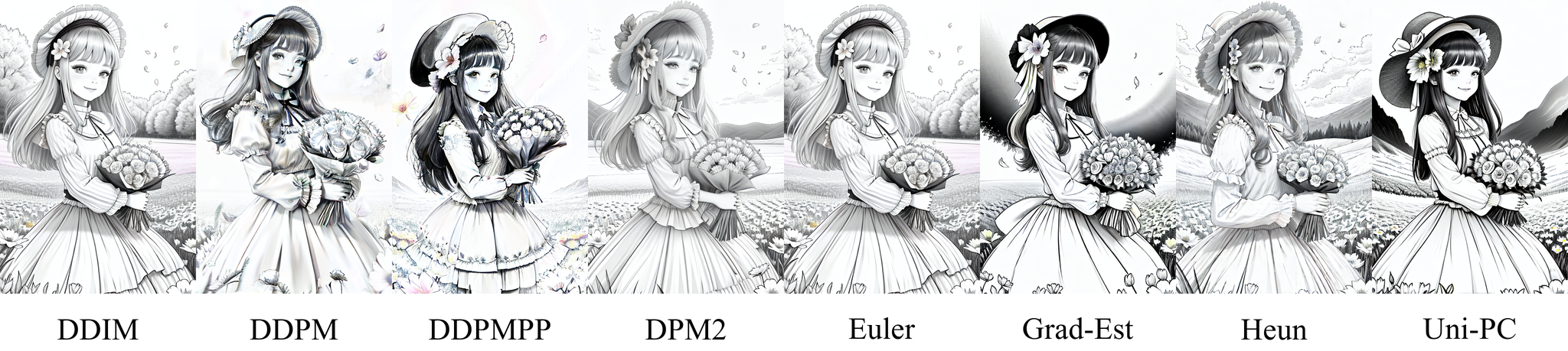}}
\caption{
\textbf{Robustness of Delta Sampling (DS) across different diffusion samplers.} 
We apply DS to transfer the \textit{animeoutlineV4\_16} LoRA (trained on SD-1.5) to SD-2.1 under a wide range of samplers, including DDIM, DDPM, DDPMPP, DPM2, Euler, Gradient Estimation Sampler (GES), Heun, and Uni-PC.  
All generations use the same prompt describing an anime-style girl in a flower field with monochrome lineart aesthetics.  
Despite substantial differences in solver dynamics, numerical stability, and stochasticity, DS consistently preserves the intended line-art adaptation across all samplers, demonstrating strong sampler-agnostic robustness.
}\label{fig:samplers}
\end{figure*}
\paragraph{Robustness Across Diverse Sampling Algorithms.}
We further assess the robustness and generality of DS by applying it under a broad suite of sampling algorithms commonly used in diffusion-based inference. In this experiment, SD-1.5 serves as the adaptation source, equipped with the \textit{animeoutlineV4\_16} LoRA\footnote{\url{https://civitai.com/models/16014/anime-lineart-manga-like-style}}, which induces a clean monochrome anime-lineart aesthetic. The target model is SD-2.1, and we evaluate whether DS can consistently transfer this stylistic effect across different samplers. We adopt the following descriptive prompt for all tests:
\begin{quote}
\small
\texttt{masterpiece, best quality, 1girl, solo, long\_hair, looking\_at\_viewer, smile, bangs, skirt, shirt, long\_sleeves, hat, dress, bow, holding, closed\_mouth, flower, frills, hair\_flower, petals, bouquet, holding\_flower, center\_frills, bonnet, holding\_bouquet, flower field, lineart, monochrome}
\end{quote}

As shown in Fig.~\ref{fig:samplers}, we test eight representative samplers: DDIM~\cite{song2021denoising}, DDPM~\cite{ho2020denoising}, DDPMPP, DPM2, Euler, the Gradient Estimation Sampler~\cite{permenter2024interpreting}, Heun~\cite{karras2022elucidating}, and Uni-PC~\cite{zhao2023unipc}. These samplers differ significantly in their discretization strategies, solver order, noise schedules, and the degree of stochasticity they introduce into the trajectory.

Across all sampling algorithms, DS consistently transfers the intended line-art and monochrome characteristics from the SD-1.5 LoRA to SD-2.1. The generated outputs reliably maintain sharp outlines, coherent edge structures, and stable stylistic cues, regardless of whether the sampler follows deterministic trajectories (e.g., DDIM, Uni-PC) or stochastic ones (e.g., DDPM, Heun). Importantly, DS requires \emph{no modification} to any sampler’s update rule: the residual-guided prediction operates entirely at the noise-estimation level and is thus orthogonal to the solver dynamics themselves.

These results highlight a key strength of DS—its compatibility with heterogeneous sampling strategies. Because DS augments predicted noise rather than altering model weights or solver equations, it integrates seamlessly into existing inference pipelines and remains effective under a wide variety of sampling regimes. This sampler-agnostic behavior underscores DS’s practical utility for real-world deployment, where inference frameworks often vary widely across applications and user preferences.



\section{Transfer Result between SD-1.5 and SD-XL}\label{sec:exp_15_xl}

\begin{figure*}[ht]
    \centering
    \includegraphics[width=0.9\linewidth]{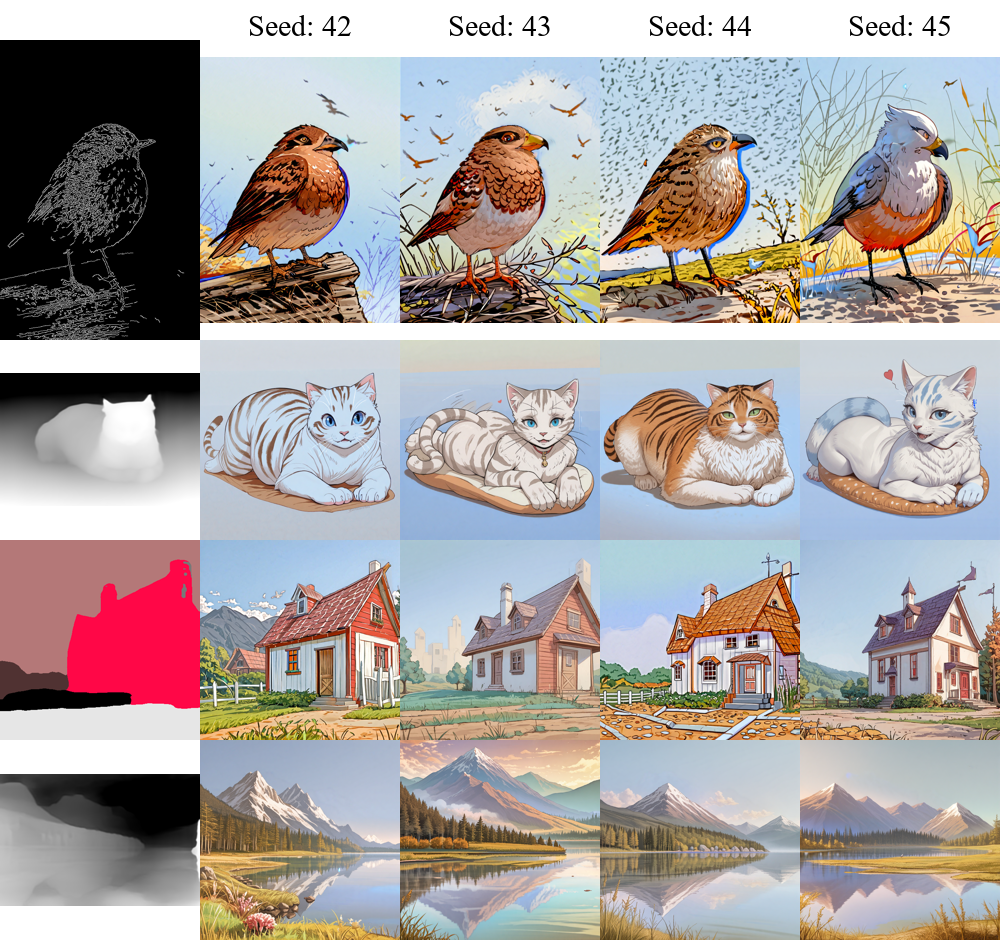}
    \caption{\textbf{Transfer of ControlNet-Based Structural Conditioning from SD-1.5 to SD-XL.}
    We evaluate DS using four different ControlNet modalities originally trained on SD-1.5: 
    (top) Canny edges of a bird, 
    (second row) depth map of a cat, 
    (third row) semantic segmentation map of a house, and 
    (bottom) depth map of a mountain lake. 
    The prompts are simply the corresponding object/scene names (``bird'', ``cat'', ``house'', ``mountain lake''). 
    From left to right, images are generated using four different seeds (42, 43, 44, 45) under identical settings.
    Across all conditions, DS enables SD-XL to faithfully preserve the structural layout encoded in the input conditioning, 
    while leveraging SD-XL’s stronger generative prior to produce images with richer details, sharper textures, and higher visual fidelity. 
    The consistent results across seeds demonstrate the stability and robustness of the proposed transfer method.
    }\label{fig:sd15-sdxl}
\end{figure*}
\paragraph{Transferring Structural Control from SD-1.5 to SD-XL.}
To further demonstrate the versatility of DS under large architectural gaps, we evaluate its performance in transferring ControlNet-based structural conditioning from SD-1.5 to SD-XL. Unlike prior experiments focused on stylistic or semantic transfer, this setup enables a more direct and visually interpretable assessment of structural consistency.

As shown in Fig.~\ref{fig:sd15-sdxl}, we adopt four representative ControlNet modalities trained on SD-1.5—\textit{Canny} (bird), \textit{depth} (cat), \textit{semantic segmentation} (house), and \textit{depth} (mountain lake). For each modality, the corresponding prompt is set to ``bird'', ``cat'', ``house'', and ``mountain lake'', respectively, while all other generation settings follow the default configuration. To evaluate robustness, we generate four samples per condition under different random seeds (\texttt{42, 43, 44, 45}). As the target backbone, we employ the widely used community fine-tuned SDXL model \texttt{prefect-pony-xl-v50-sdxl}.\footnote{\url{https://huggingface.co/John6666/prefect-pony-xl-v50-sdxl}}

Across all conditions and seeds, DS successfully transfers the structural constraints imposed by the SD-1.5 ControlNet, enabling SD-XL to faithfully preserve key geometric features such as contour, layout, and shape. At the same time, the SD-XL backbone contributes significantly enhanced detail, richer textures, vivid color rendering, and globally coherent high-resolution structure. The consistency observed across different seeds indicates that DS produces stable and reliable guidance, effectively combining SD-1.5 control signals with the superior generative capacity of SD-XL. The consistency of the outputs across different seeds demonstrates the stability and reliability of DS.

\paragraph{Reverse Transfer: From SD-XL to SD-1.5.}
We further examine the reverse transfer setting, where the adaptation originates from SD-XL while the target model is SD-1.5. Unlike SD-1.5, for which a rich variety of community-produced ControlNet modules exist, SD-XL ControlNet models remain extremely limited. In practice, we were only able to locate two SD-XL ControlNet variants---\textit{Canny} and \textit{depth}---from the collection\footnote{\url{https://huggingface.co/lllyasviel/sd_control_collection}}. Consequently, we focus our evaluation on these two modalities.

\begin{figure*}[ht]
    \centering
    \includegraphics[width=0.8\linewidth]{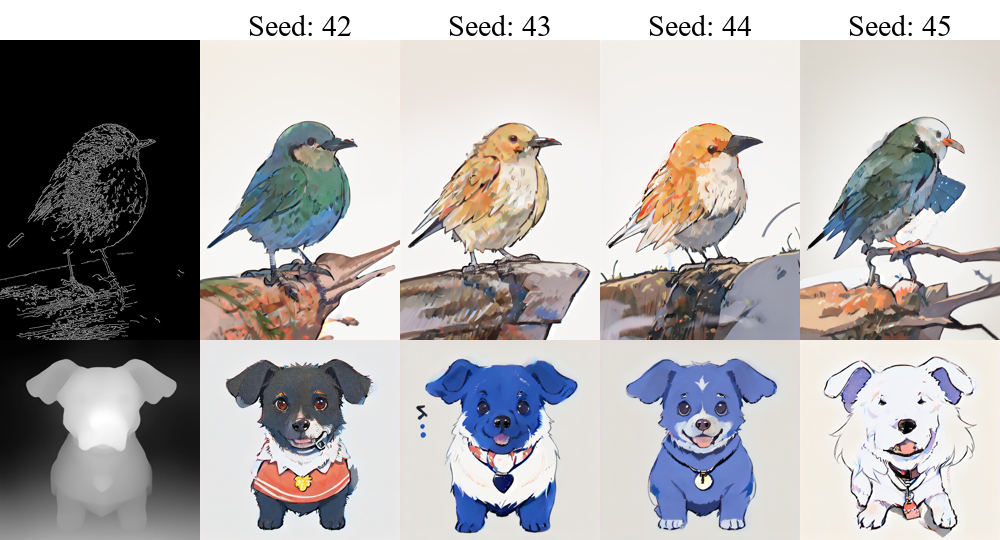}
    \caption{
    \textbf{Reverse transfer from SD-XL ControlNet to SD-1.5 using Delta Sampling.}
    Due to the scarcity of SD-XL ControlNet models, only two modalities are available: 
    (1) \textit{Canny} and (2) \textit{depth}.\ 
    The first row shows results using a SD-XL \textit{Canny} ControlNet conditioned on a bird image, while the second row shows results using a SD-XL \textit{depth} ControlNet conditioned on a dog image. 
    For each condition, four samples are generated using seeds 42, 43, 44, and 45 (left to right). 
    DS enables SD-1.5 to accurately preserve the structural constraints imposed by the SD-XL ControlNet despite substantial architectural differences. 
    Although the overall fidelity and detail are limited by the capacity of SD-1.5, the transferred structural guidance remains stable and consistent across different seeds. 
    }
    \label{fig:sdxl-sd15}
\end{figure*}

As shown in Fig.~\ref{fig:sdxl-sd15}, the first row illustrates results produced using a SD-XL \textit{Canny} ControlNet conditioned on a bird image, while the second row shows transfer from a SD-XL \textit{depth} ControlNet conditioned on a dog image. For each condition, we generate outputs using four random seeds (\texttt{42, 43, 44, 45}) to test stability. Across both modalities, DS enables SD-1.5 to faithfully follow the structural constraints imposed by the SD-XL ControlNet, demonstrating that the residual signal effectively captures the cross-model control behavior even in the reverse direction.

Compared with the earlier SD-1.5~$\rightarrow$~SD-XL results, the images generated by SD-1.5 exhibit relatively weaker detail, sharpness, and texture richness due to the inherent capacity limitations of SD-1.5. Nevertheless, the structural alignment remains strong and consistent across seeds, underscoring the robustness of DS.

Importantly, the scarcity of SD-XL ControlNet models highlights the practical necessity of our approach. DS allows practitioners to leverage the abundant SD-1.5 ControlNet ecosystem to achieve precise structural control on SD-XL, without the need to train costly SD-XL-specific ControlNets. This capability significantly lowers the barrier for enabling complex controlled generation on modern high-capacity diffusion models.

\section{Transfer result between SD-3 and SD-3.5}\label{sec:exp_3_35}
\paragraph{Cross-Generation Structural Transfer Between SD-3 and SD-3.5.}
We further evaluate DS in the context of the recent SD-3 and SD-3.5 model families. Compared with earlier generations such as SD-1.5 and SD-XL, these backbones exhibit substantially stronger visual fidelity, more advanced semantic reasoning, and higher-resolution capabilities. However, the ecosystem of ControlNet models for SD-3 and SD-3.5 remains limited. For SD-3, only three community-trained ControlNets are currently available—\textit{Pose}, \textit{Canny}, and \textit{Tile}—all released by InstantX\footnote{\url{https://huggingface.co/InstantX/SD3-Controlnet-Canny}}. For SD-3.5, the official collection from Stability AI\footnote{\url{https://huggingface.co/stabilityai/stable-diffusion-3.5-controlnets}} provides three modalities (\textit{blur}, \textit{canny}, and \textit{depth}), supporting SD-3.5-Large only.

\begin{figure*}[ht]
    \centering
    \includegraphics[width=\linewidth]{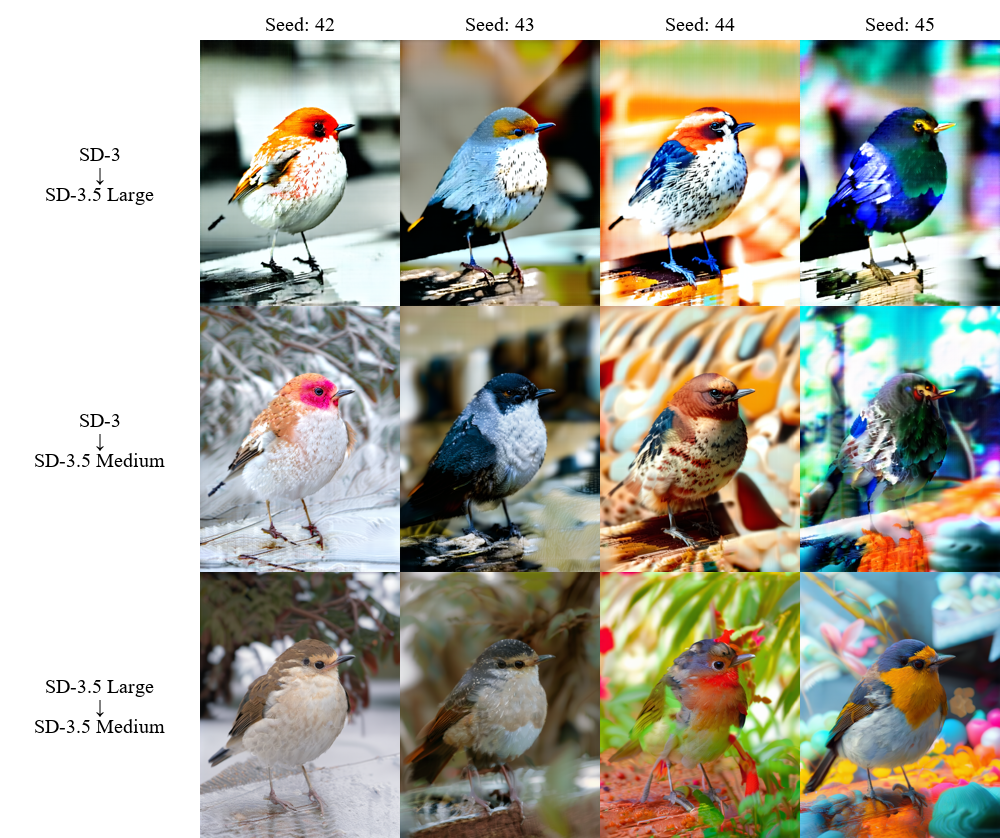}
    \caption{
    \textbf{Cross-generation ControlNet transfer between SD-3 and SD-3.5 using Delta Sampling.}
    Due to limited ControlNet availability for these models, we use the \textit{Canny} modality for both SD-3 (InstantX) and SD-3.5-Large (Stability AI). 
    A canny bird image serves as the conditioning input for all experiments. 
    From top to bottom, we present results from three transfer directions:
    (1) SD-3~$\rightarrow$~SD-3.5-Large,
    (2) SD-3~$\rightarrow$~SD-3.5-Medium, and
    (3) SD-3.5-Large~$\rightarrow$~SD-3.5-Medium.
    For each direction, four images are generated using seeds 42, 43, 44, and 45 (left to right). 
    DS enables the target models to accurately preserve structural constraints while leveraging the strong generative capability of SD-3 and SD-3.5, resulting in images with highly consistent structure and rich fine-grained detail. 
    These results highlight the robustness of DS and its ability to facilitate structural control transfer even between high-capacity, cross-generation diffusion models.
    }
    \label{fig:sd3-sd35}
\end{figure*}

To enable consistent comparison across model variants, we use the \textit{Canny} modality for both SD-3 and SD-3.5-Large. A bird image is used as the conditioning input for all experiments. As shown in Fig.~\ref{fig:sd3-sd35}, we evaluate three transfer directions: 
(i) SD-3~$\rightarrow$~SD-3.5-Large, 
(ii) SD-3~$\rightarrow$~SD-3.5-Medium, and 
(iii) SD-3.5-Large~$\rightarrow$~SD-3.5-Medium. 
For each case, we generate four samples under different random seeds (\texttt{42, 43, 44, 45}). All experiments are performed at a high resolution of $768 \times 1024$.

Across all transfer directions, DS successfully preserves the structural constraints imposed by the canny conditioning, enabling the target models to maintain accurate contour information and spatial arrangement. Meanwhile, the SD-3 and SD-3.5 backbones contribute high-quality rendering with rich textures, fine detail, and strong color fidelity. The consistency of results across seeds illustrates the robustness of DS even under cross-generation transfer between closely related but architecturally distinct high-capacity diffusion models.

These findings show that DS effectively bridges the feature-space gap between SD-3 and SD-3.5, enabling flexible reuse of structural control modules across model variants and underscoring its practical value in scenarios where ControlNet availability is sparse.

\section{IP-Adapter Transfer result}\label{sec:exp_ip}
\begin{figure*}[ht]
    \centering
    \includegraphics[width=\linewidth]{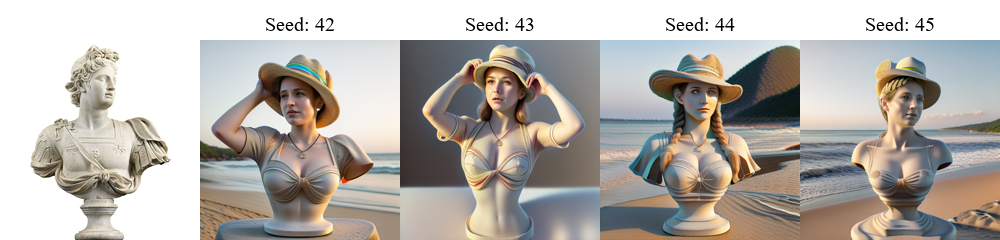}
    \caption{
    \textbf{Transferring IP-Adapter conditioning from SD-1.5 to SD-2.1 using Delta Sampling.}
    We use the canonical \textit{statue} reference image from the IP-Adapter paper~\cite{ye2023ip}(left) and generate outputs with the prompt 
    \texttt{``best quality, high quality, wearing a hat on the beach''}. 
    The four images on the right are produced using seeds 42, 43, 44, and 45 (left to right). 
    DS successfully preserves the structural identity of the statue while integrating new contextual attributes such as the beach environment and the hat described in the prompt. 
    The consistent results across seeds demonstrate that DS enables robust transfer of image-based conditioning effects from IP-Adapter, even when the target model differs architecturally from the original training backbone.
    }
    \label{fig:ip-adapter}
\end{figure*}
\paragraph{Transferring Image-Based Conditioning via IP-Adapter.}
Beyond parameter-efficient fine-tuning and structural guidance, we further evaluate DS on transferring image-based conditioning signals using IP-Adapter~\cite{ye2023ip}. Specifically, we use an IP-Adapter model originally trained for SD-1.5 and test whether its conditioning behavior can be transferred to SD-2.1 without any retraining. Following the evaluation protocol in the original IP-Adapter paper, we adopt the classic \textit{statue} example as the reference image (shown on the left of Fig.~\ref{fig:ip-adapter}). The generation prompt is set to \texttt{``best quality, high quality, wearing a hat on the beach''}. To assess stability, we generate four samples under different random seeds (\texttt{42, 43, 44, 45}).

As illustrated in Fig.~\ref{fig:ip-adapter}, DS successfully preserves the structural characteristics of the statue while enabling SD-2.1 to reinterpret the subject within a new scene, faithfully rendering the ``beach'' environment and the additional ``hat'' attribute described by the prompt. The consistency across seeds further indicates that DS can robustly transfer high-level appearance and identity cues from IP-Adapter, even when applied to a target model with a different architecture. These results demonstrate that DS is not limited to text- or structure-based adaptations but also effectively supports image-based conditioning, substantially broadening its applicability within the diffusion model ecosystem.

\section{Prompt Engineering Transfer Result}\label{sec:exp_prompt}

\begin{figure*}[ht]
    \centering
    \includegraphics[width=\linewidth]{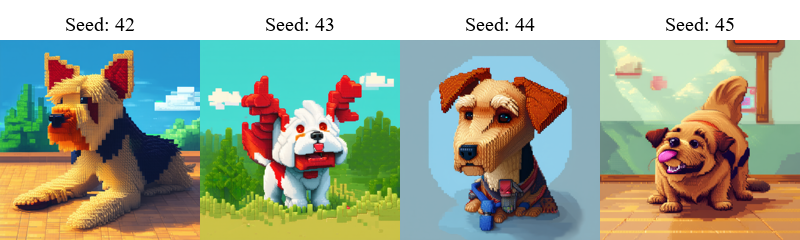}
    \caption{
    \textbf{Prompt-engineering transfer from SD-3.5-Medium to SD-3.5-Large using Delta Sampling.}
    A LoRA fine-tuned on SD-3.5-Medium for pixel-art generation is applied to the base prompt 
    \texttt{``Pixel Art, dog''}, while the SD-3.5-Large target model receives only the prompt 
    \texttt{``dog''}. 
    Four samples (seeds 42, 43, 44, 45, left to right) are generated at $1024 \times 1024$ resolution. 
    Despite the absence of the ``Pixel Art'' keyword in the target prompt, DS successfully transfers the pixel-art aesthetic to SD-3.5-Large, producing outputs with clear pixel-art stylization while maintaining semantic fidelity. 
    This demonstrates that DS effectively transfers prompt-induced stylistic modulation across models without requiring prompt alignment.
    }
    \label{fig:prompt-transfer}
\end{figure*}
\paragraph{Transferring Prompt-Driven Style Modulation via Prompt Engineering.}
We evaluate whether DS can transfer prompt-driven stylistic behavior that arises from prompt engineering or prompt-specific fine-tuning. In this experiment, we use SD-3.5-Medium as the base model and apply a LoRA fine-tuned for pixel-art generation.\footnote{\url{https://civitai.com/models/144684/pixelartredmond-pixel-art-loras-for-sd-xl}} 
This LoRA is specifically optimized to induce a ``pixel art'' aesthetic when used with prompts containing corresponding style descriptors. Our goal is to determine whether the pixel-art effect learned on SD-3.5-Medium can be transferred to SD-3.5-Large, \emph{even when the target prompt does not contain any pixel-art keywords}. 

To this end, we use the prompt \texttt{``Pixel Art, dog''} for the adapted SD-3.5-Medium model, while the SD-3.5-Large target model receives only the simplified prompt \texttt{``dog''}. All images are generated at a resolution of $1024 \times 1024$, and four seeds (\texttt{42, 43, 44, 45}) are used to assess consistency.

As shown in Fig.~\ref{fig:prompt-transfer}, DS successfully transfers the stylistic effect induced by the pixel-art LoRA from SD-3.5-Medium to SD-3.5-Large. Despite the absence of explicit pixel-art descriptors in the target prompt, the resulting images clearly exhibit strong pixel-art characteristics while maintaining the semantic content of the prompt. This demonstrates that DS can propagate prompt-driven or prompt-conditioned stylistic biases across models, enabling powerful cross-model reuse of prompt-engineering effects without requiring the target model to include the same stylistic cues in its prompt.

\section{Multiple Adapation Transfer Result}\label{sec:exp_mult_adapt}
\begin{figure*}[ht]
\centering
\centerline{\includegraphics[width=\linewidth]{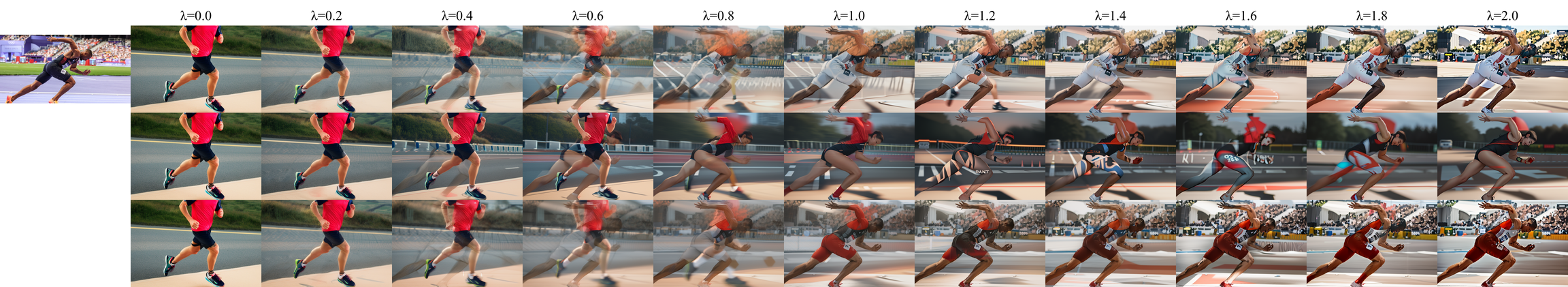}}
\caption{
\textbf{Qualitative results of transferring composed ControlNet conditions using Delta Sampling (DS).} 
Columns vary the guidance strength ($\lambda \in \{0.0, 0.2, \dots, 2.0\}$), while rows compare three scenarios: 
\textbf{(Top)} applying a single \textit{Depth} ControlNet, 
\textbf{(Middle)} applying a single \textit{Canny} ControlNet, and 
\textbf{(Bottom)} applying a composite configuration that merges both Depth and Canny ControlNets at a 1:1 ratio. 
The combined case highlights DS's ability to accurately transfer mixed structural constraints and to preserve the joint effects of multiple heterogeneous conditioning signals.
}
\label{fig:multicnet}
\end{figure*}
\paragraph{Composing Multiple Structural Adaptations.}
To further evaluate the flexibility of DS, we investigate its ability to handle multiple ControlNet modules applied simultaneously. This setting is particularly important in real-world workflows, where users often combine several structural constraints—such as depth, edges, pose, or segmentation—to achieve fine-grained control over image geometry. Unlike weight-space transfer methods, which typically require architectural compatibility and explicit module alignment, DS naturally accommodates such compositions because the residual $\Delta$ inherently captures the aggregate prediction behavior of all active modules.

As shown in Fig.~\ref{fig:multicnet}, we consider three configurations: 
(i) a single \textit{Depth} ControlNet, 
(ii) a single \textit{Canny} ControlNet, and 
(iii) a composite configuration that blends Depth and Canny ControlNets in equal proportion. 
For each configuration, we sweep the guidance strength $\lambda$ from 0.0 to 2.0. 
The single-module settings illustrate that DS faithfully transfers each individual structural constraint, reproducing the expected depth-aware geometry and edge-aligned contours on the target model. 
More importantly, the composite case demonstrates that DS preserves the \emph{combined} structural influence, producing outputs that simultaneously satisfy depth consistency and sharp edge alignment. 
This indicates that DS treats the residual signal as a unified control representation, enabling it to transfer complex, multi-source structural adaptations without requiring any retraining or module-specific engineering.

These results highlight DS's practical utility in scenarios involving modular and compositional conditioning, underscoring its compatibility with the highly flexible, community-driven ControlNet ecosystem.

\begin{figure*}[ht]
\centering
\centerline{\includegraphics[width=\linewidth]{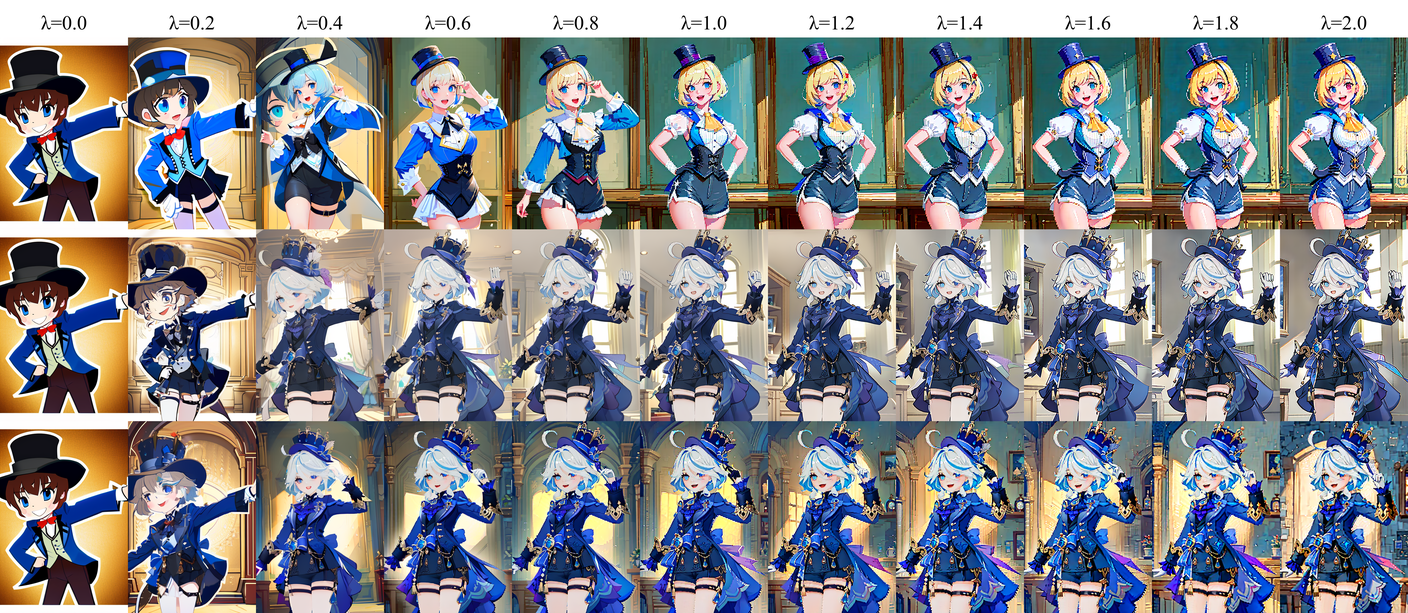}}
\caption{
\textbf{Qualitative results of transferring composed LoRA adaptations using Delta Sampling (DS).} 
Columns vary the guidance strength ($\lambda \in \{0.0, 0.2, \dots, 2.0\}$), while rows illustrate three scenarios: 
\textbf{(Top)} applying a single \textit{Pixel Art} LoRA, 
\textbf{(Middle)} applying a single character-specific \textit{Furina} LoRA, and 
\textbf{(Bottom)} applying a composite configuration that merges both LoRAs at a 1:1 ratio. 
The combined case demonstrates DS’s ability to faithfully transfer mixed stylistic and identity-specific adaptations, preserving the joint influence of multiple LoRA modules without requiring any retraining or weight merging.
}
\label{fig:multilora}

\end{figure*}
\paragraph{Composing Multiple LoRA Adaptations.}
In addition to structural modules such as ControlNet, many real-world diffusion workflows rely heavily on the composition of multiple LoRA adapters. Users frequently combine style-oriented LoRAs with character- or object-specific LoRAs to achieve complex and fine-grained control over appearance. Traditional merging strategies often require manual weight fusion or careful hyperparameter tuning, and failures can occur when LoRAs interact in non-linear or incompatible ways. In contrast, DS avoids these issues entirely: because it operates in prediction space, the residual signal automatically captures the combined influence of all active LoRA modules.

Fig.~\ref{fig:multilora} evaluates three such scenarios:  
(i) a \textit{Pixel Art} LoRA that induces a coarse, blocky aesthetic;  
(ii) a character-specific \textit{Furina} LoRA capturing identity and facial attributes; and  
(iii) a composite 1:1 mixture of both LoRAs.  
For each configuration, we sweep the guidance strength $\lambda$ from 0.0 to 2.0. The single-LoRA cases confirm that DS faithfully transfers each individual adaptation, reproducing the expected stylistic or identity-specific effects on the target model. In the composite setting, DS successfully preserves the joint influence of both LoRAs, producing images that exhibit the pixel-art aesthetic while also retaining the character-specific features introduced by the Furina LoRA.

These results demonstrate that DS seamlessly supports multi-LoRA composition, enabling flexible reuse of style and identity modules across different diffusion backbones. Unlike weight merging or architecture-specific fusion techniques, DS requires no additional training or LoRA-specific adjustments, making it a practical and robust solution for real-world composition workflows.

\begin{figure*}[ht]
\centering
\centerline{\includegraphics[width=\linewidth]{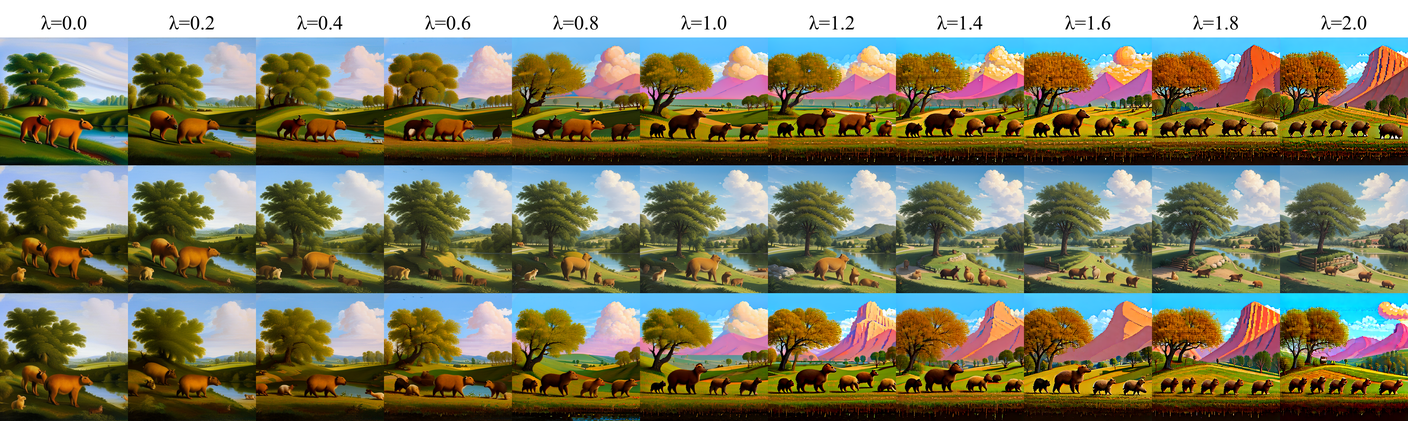}}
\caption{
\textbf{Qualitative results of cross-model LoRA composition using Delta Sampling (DS).} 
Columns vary the guidance strength ($\lambda \in \{0.0, 0.2, \dots, 2.0\}$). 
Rows illustrate three scenarios: 
\textbf{(Top)} applying a \textit{Pixel Art} LoRA originally trained on SD-1.5, 
\textbf{(Middle)} applying a \textit{Nicolas Poussin} artistic-style LoRA natively trained on SD-2.1, and 
\textbf{(Bottom)} applying a composite configuration that merges both LoRAs (SD-1.5 Pixel Art + SD-2.1 Poussin) at a 1:1 ratio. 
The combined case demonstrates DS’s ability to unify adaptations originating from different base architectures, enabling coherent cross-model stylistic composition without retraining or weight merging.
}
\label{fig:multiloradiffbase}
\end{figure*}
\paragraph{Cross-Model Composition of LoRA Adaptations.}
Beyond composing multiple LoRA modules trained on the same backbone, a more challenging and practically relevant scenario arises when the LoRAs originate from \emph{different diffusion architectures}. This situation frequently occurs in community-driven ecosystems such as SD-1.x, SD-2.x, SD-XL, and SD-3.x, where stylistic LoRAs trained on earlier models are not directly compatible with newer architectures. Traditional merging or conversion methods often fail due to differences in parameterization, attention block layout, or channel dimensions, making cross-model LoRA reuse extremely difficult.

DS overcomes this barrier naturally. Since the residual $\Delta$ reflects the behavioral effect of each LoRA in prediction space, it remains portable even when the underlying architectures differ. To evaluate this capability, we consider three configurations shown in Fig.~\ref{fig:multiloradiffbase}:  
(i) a \textit{Pixel Art} LoRA trained on SD-1.5,  
(ii) a \textit{Nicolas Poussin} style LoRA natively trained on SD-2.1, and  
(iii) a cross-model composite where both LoRAs are applied simultaneously.  
For each case, we vary the guidance strength $\lambda$ from 0.0 to 2.0.

The single-LoRA results confirm that DS successfully transfers stylistic effects from both architectures: the SD-1.5 Pixel Art LoRA induces its distinct blocky aesthetic, while the SD-2.1 Poussin LoRA produces painterly, classical textures. Remarkably, the composite scenario demonstrates that DS can \emph{combine} stylistic signals from incompatible models, generating images that exhibit both pixel-art characteristics and Poussin-inspired brushwork in a coherent manner. This indicates that DS effectively disentangles weight compatibility from stylistic behavior, enabling cross-backbone LoRA composition without any retraining or architectural alignment.

These findings highlight the unique flexibility of DS in enabling cross-model reuse and combination of stylistic modules, opening the door to seamless integration of community-created LoRA assets across heterogeneous diffusion architectures.

\section{More Quantitative Results on LoRA and LoHa.}\label{sec:exp_more_lyco}
\paragraph{Similarity--Diversity Trade-off Across Categories.}
We further analyze the behavior of DS across five representative semantic categories: \textit{Scene}, \textit{Anime}, \textit{People}, \textit{Style}, and \textit{Toy}. For each category, we report (i) \emph{similarity} between generated images and the target LoRA/LoHa adaptation (e.g., CLIP-based similarity to reference exemplars), and (ii) \emph{diversity} among generated outputs for the same prompt (e.g., LPIPS across image pairs). All experiments use the same configuration described in Sec.~X, enabling consistent comparison across categories.

\begin{figure*}[ht]
   \centering
   \begin{subfigure}[t]{0.9\linewidth}
     \centering
    \includegraphics[width=\linewidth]{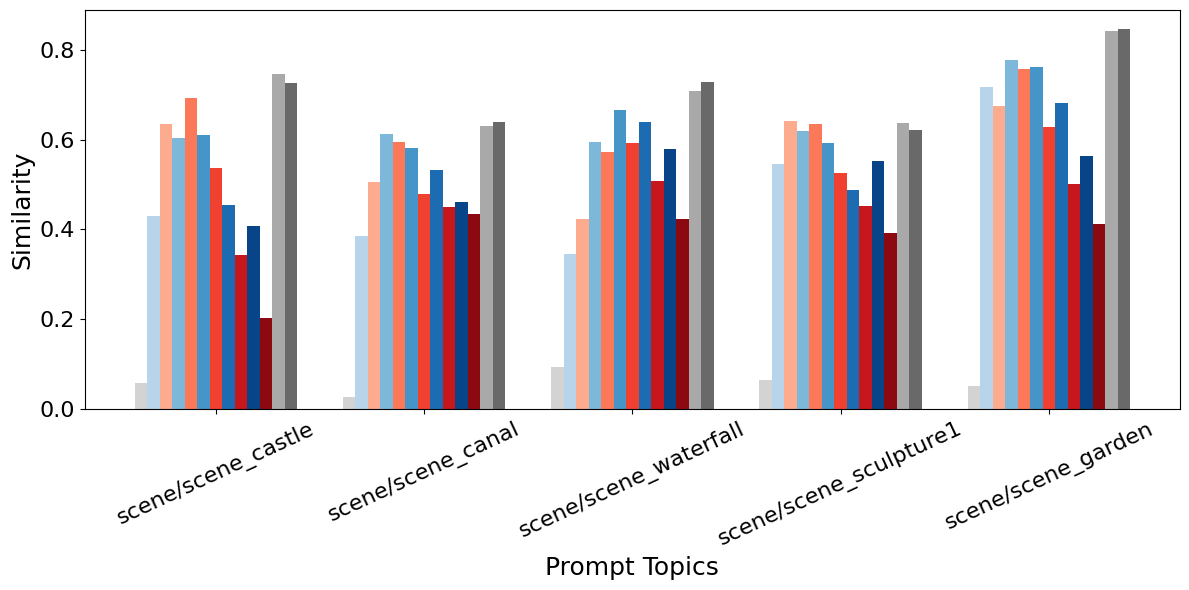}
    \caption{Similarity of category "Scene"}
    \label{fig:sub1}
  \end{subfigure}
  \begin{subfigure}[t]{0.9\linewidth}
    \centering
    \includegraphics[width=\linewidth]{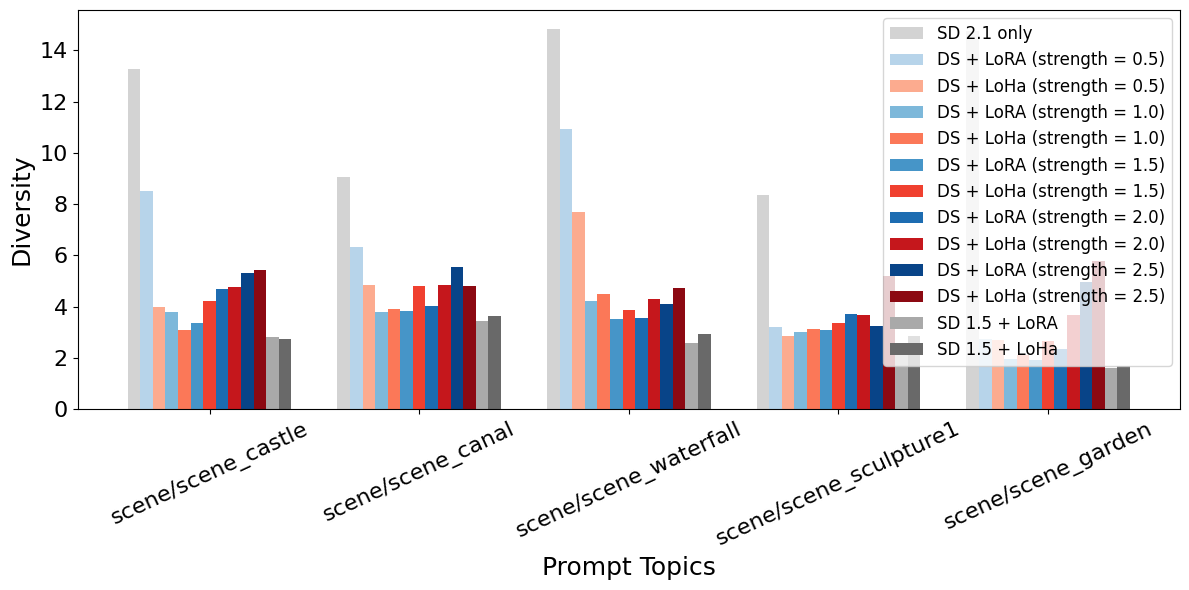}
    \caption{Diversity of category "Scene"}
    \label{fig:sub2}
  \end{subfigure}
    \caption{We established three baselines: (1) SD-2.1 Only (no adaptation), (2) SD-1.5 with the original 32-rank, 16-alpha LoRA (target effect with LoRA), and (3) SD-1.5 with the original 16-rank, 8-alpha LoHa (target effect with LoHa). In comparison, we swept the guidance strength \( \lambda \in \{0.5, 1.0, \dots, 2.5\} \) for both (1) DS applying a 32-rank, 16-alpha LoRA (DS + LoRA) to SD-2.1, and (2) DS applying a 16-rank, 8-alpha LoHa (DS + LoHa) to SD-2.1. (a) Similarity measures adherence to the target style/concept (e.g., using CLIP-Score with reference prompts/images). (b) Diversity assesses visual variation among images generated for the same prompt (e.g., using average LPIPS between pairs).}
    \label{fig:combined_scene}
\end{figure*}

\begin{figure*}[ht]
   \centering
   \begin{subfigure}[t]{0.9\linewidth}
     \centering
    \includegraphics[width=\linewidth]{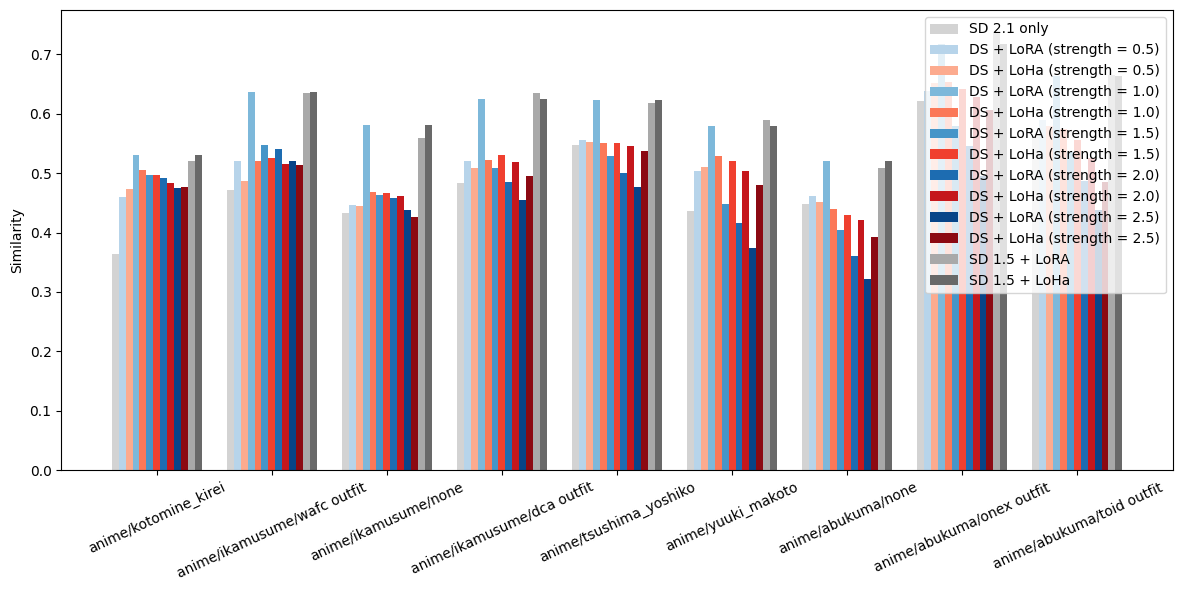}
    \caption{Similarity of category "Anime"}
    \label{fig:sub3}
  \end{subfigure}
  \begin{subfigure}[t]{0.9\linewidth}
    \centering
    \includegraphics[width=\linewidth]{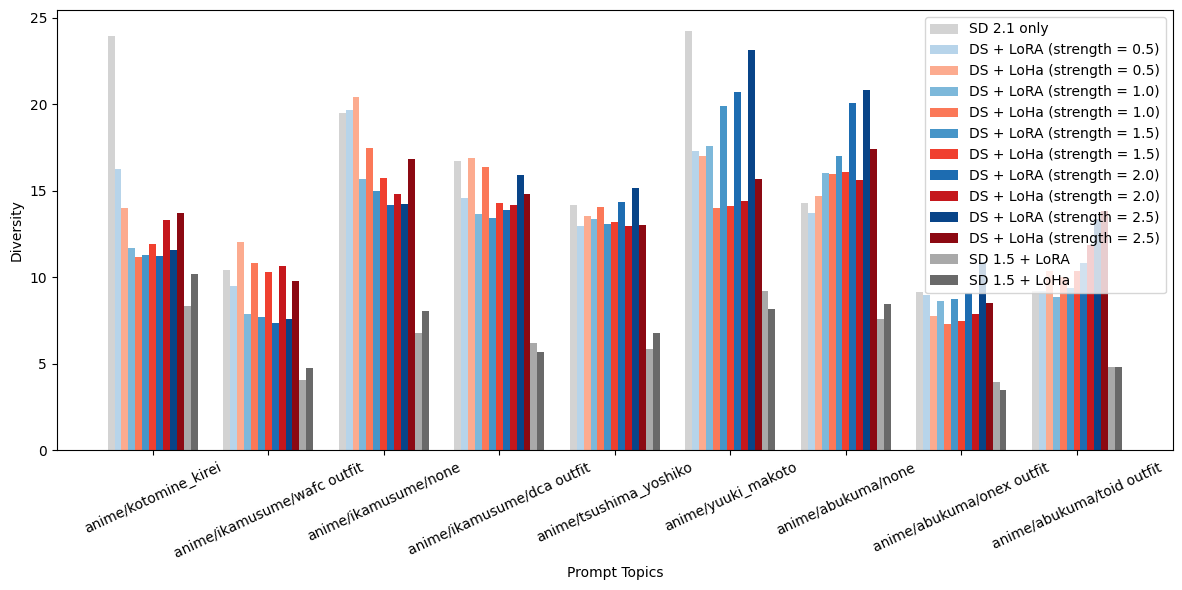}
    \caption{Diversity of category "Anime"}
    \label{fig:sub4}
  \end{subfigure}
    \caption{We established three baselines: (1) SD-2.1 Only (no adaptation), (2) SD-1.5 with the original 32-rank, 16-alpha LoRA (target effect with LoRA), and (3) SD-1.5 with the original 16-rank, 8-alpha LoHa (target effect with LoHa). In comparison, we swept the guidance strength \( \lambda \in \{0.5, 1.0, \dots, 2.5\} \) for both (1) DS applying a 32-rank, 16-alpha LoRA (DS + LoRA) to SD-2.1, and (2) DS applying a 16-rank, 8-alpha LoHa (DS + LoHa) to SD-2.1. (a) Similarity measures adherence to the target style/concept (e.g., using CLIP-Score with reference prompts/images). (b) Diversity assesses visual variation among images generated for the same prompt (e.g., using average LPIPS between pairs).}
    \label{fig:combined_anime}
\end{figure*}

\begin{figure*}[ht]
   \centering
   \begin{subfigure}[t]{0.9\linewidth}
     \centering
    \includegraphics[width=\linewidth]{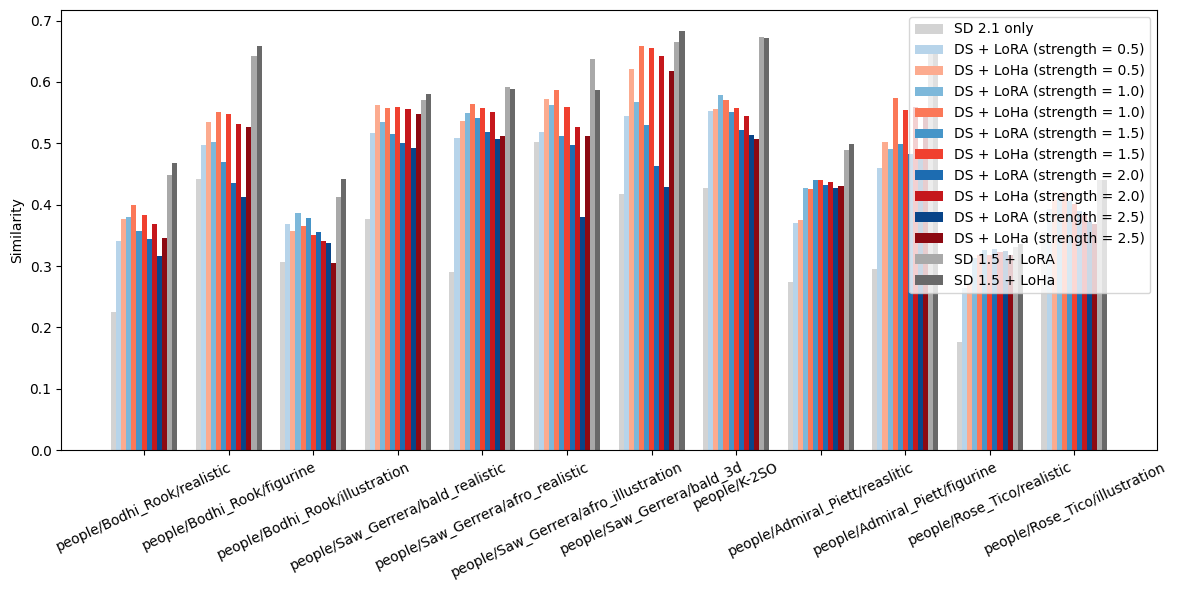}
    \caption{Similarity of category "People"}
    \label{fig:sub5}
  \end{subfigure}
  \begin{subfigure}[t]{0.9\linewidth}
    \centering
    \includegraphics[width=\linewidth]{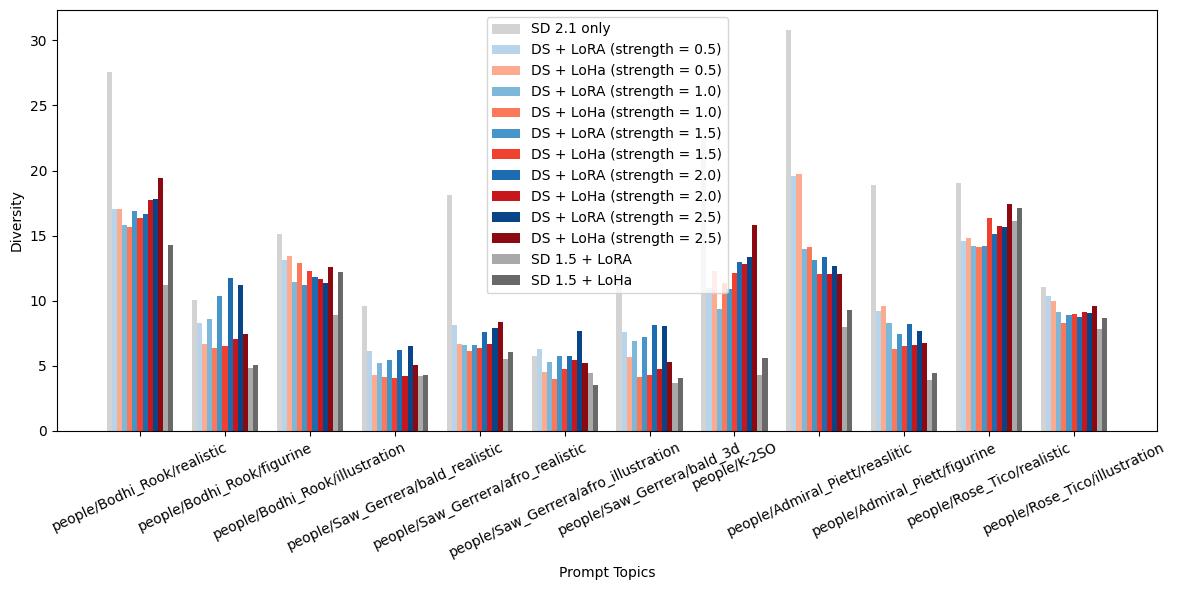}
    \caption{Diversity of category "People"}
    \label{fig:sub6}
  \end{subfigure}
    \caption{We established three baselines: (1) SD-2.1 Only (no adaptation), (2) SD-1.5 with the original 32-rank, 16-alpha LoRA (target effect with LoRA), and (3) SD-1.5 with the original 16-rank, 8-alpha LoHa (target effect with LoHa). In comparison, we swept the guidance strength \( \lambda \in \{0.5, 1.0, \dots, 2.5\} \) for both (1) DS applying a 32-rank, 16-alpha LoRA (DS + LoRA) to SD-2.1, and (2) DS applying a 16-rank, 8-alpha LoHa (DS + LoHa) to SD-2.1. (a) Similarity measures adherence to the target style/concept (e.g., using CLIP-Score with reference prompts/images). (b) Diversity assesses visual variation among images generated for the same prompt (e.g., using average LPIPS between pairs).}
    \label{fig:combined_people}
\end{figure*}

\begin{figure*}[ht]
   \centering
   \begin{subfigure}[t]{0.9\linewidth}
     \centering
    \includegraphics[width=\linewidth]{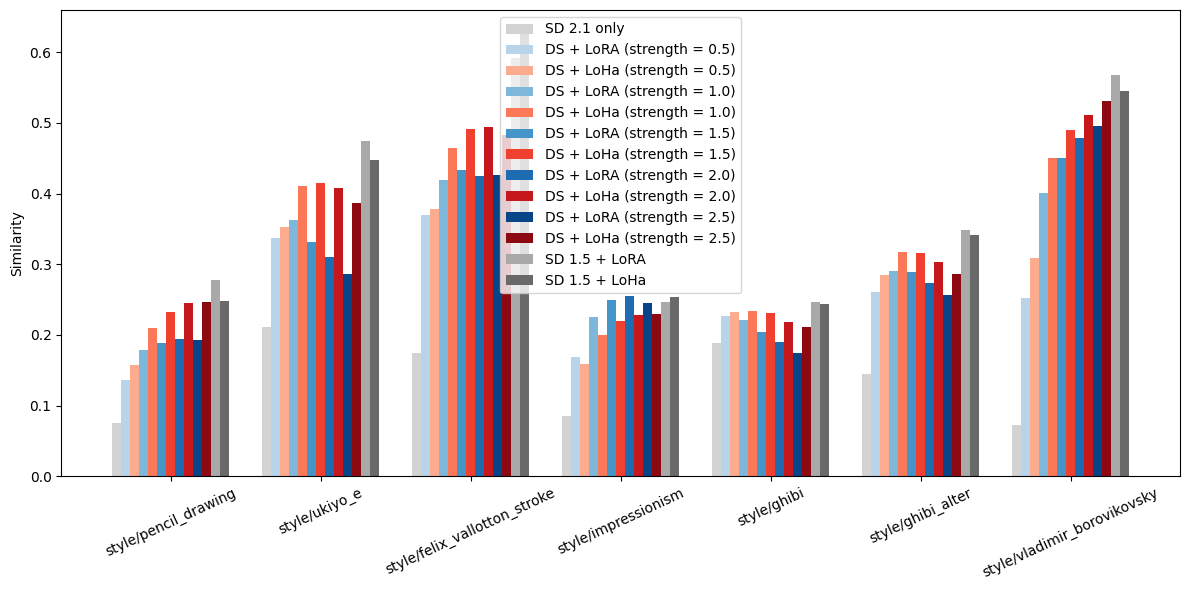}
    \caption{Similarity of category "Style"}
    \label{fig:sub7}
  \end{subfigure}
  \begin{subfigure}[t]{0.9\linewidth}
    \centering
    \includegraphics[width=\linewidth]{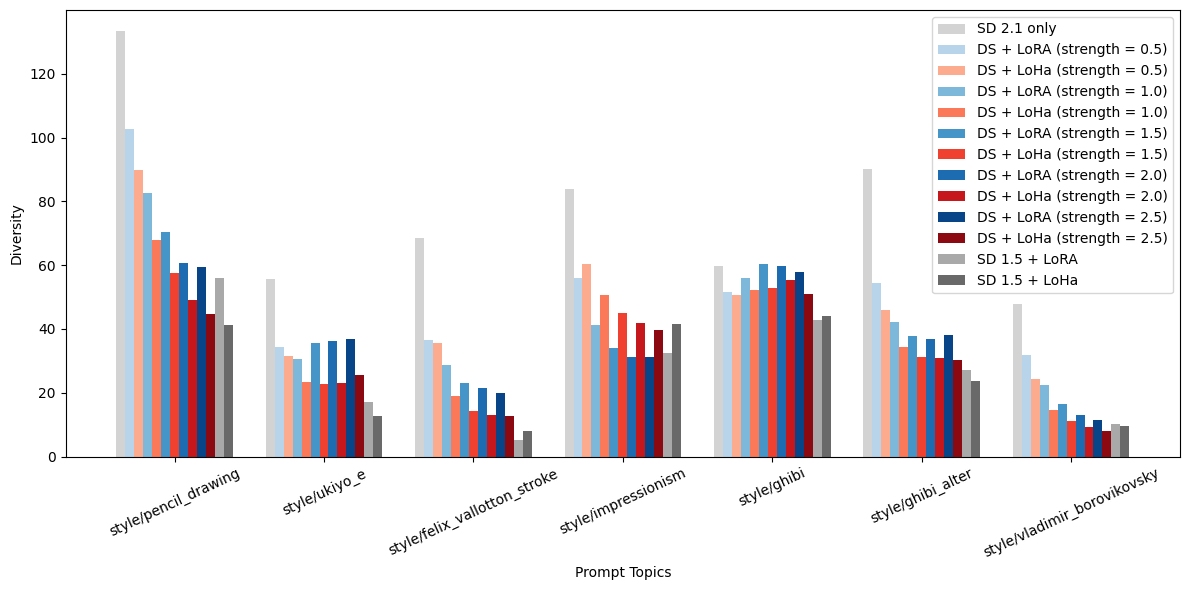}
    \caption{Diversity of category "Style"}
    \label{fig:sub8}
  \end{subfigure}
    \caption{We established three baselines: (1) SD-2.1 Only (no adaptation), (2) SD-1.5 with the original 32-rank, 16-alpha LoRA (target effect with LoRA), and (3) SD-1.5 with the original 16-rank, 8-alpha LoHa (target effect with LoHa). In comparison, we swept the guidance strength \( \lambda \in \{0.5, 1.0, \dots, 2.5\} \) for both (1) DS applying a 32-rank, 16-alpha LoRA (DS + LoRA) to SD-2.1, and (2) DS applying a 16-rank, 8-alpha LoHa (DS + LoHa) to SD-2.1. (a) Similarity measures adherence to the target style/concept (e.g., using CLIP-Score with reference prompts/images). (b) Diversity assesses visual variation among images generated for the same prompt (e.g., using average LPIPS between pairs).}
    \label{fig:combined_style}
\end{figure*}

\begin{figure*}[ht]
   \centering
   \begin{subfigure}[t]{0.9\linewidth}
     \centering
    \includegraphics[width=\linewidth]{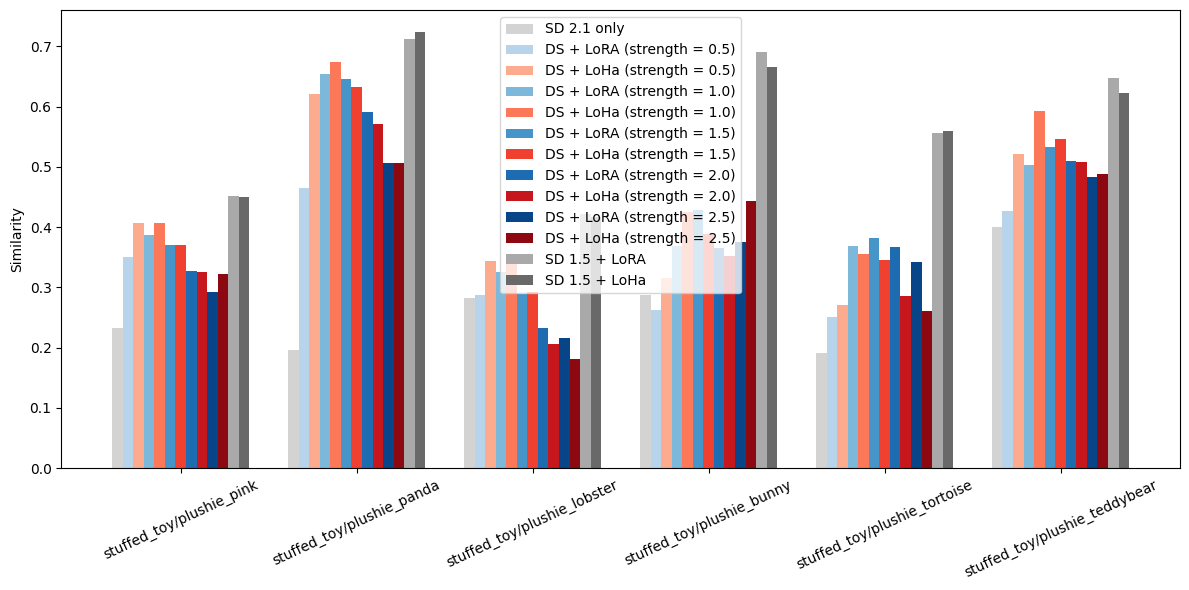}
    \caption{Similarity of category "Toy"}
    \label{fig:sub9}
  \end{subfigure}
  \begin{subfigure}[t]{0.9\linewidth}
    \centering
    \includegraphics[width=\linewidth]{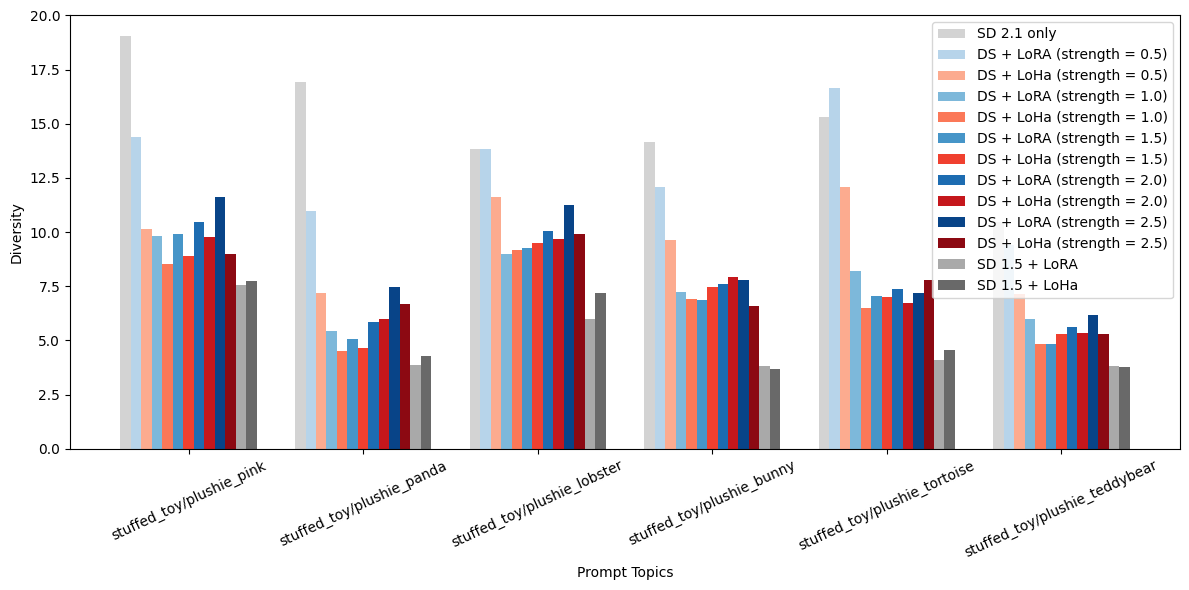}
    \caption{Diversity of category "Toy"}
    \label{fig:sub10}
  \end{subfigure}
    \caption{We established three baselines: (1) SD-2.1 Only (no adaptation), (2) SD-1.5 with the original 32-rank, 16-alpha LoRA (target effect with LoRA), and (3) SD-1.5 with the original 16-rank, 8-alpha LoHa (target effect with LoHa). In comparison, we swept the guidance strength \( \lambda \in \{0.5, 1.0, \dots, 2.5\} \) for both (1) DS applying a 32-rank, 16-alpha LoRA (DS + LoRA) to SD-2.1, and (2) DS applying a 16-rank, 8-alpha LoHa (DS + LoHa) to SD-2.1. (a) Similarity measures adherence to the target style/concept (e.g., using CLIP-Score with reference prompts/images). (b) Diversity assesses visual variation among images generated for the same prompt (e.g., using average LPIPS between pairs).}
    \label{fig:combined_toy}
\end{figure*}

Fig.~\ref{fig:combined_scene} shows the results for the \textit{Scene} category, while analogous results for \textit{Anime}, \textit{People}, \textit{Style}, and \textit{Toy} are provided in Fig.~\ref{fig:combined_anime}, Fig.~\ref{fig:combined_people}, Fig.~\ref{fig:combined_style}, and Fig.~\ref{fig:combined_toy}, respectively. Despite differences in visual domain and adaptation strength, all categories exhibit a remarkably consistent trend.

First, the \emph{similarity} curves follow a characteristic rise--fall pattern as $\lambda$ increases. For small guidance strengths, similarity increases steadily as the residual signal injects more stylistic or semantic information from the SD-1.5 LoRA/LoHa modules into SD-2.1. This corresponds to DS effectively amplifying the target adaptation. As $\lambda$ approaches a moderate value (typically between $1.0$ and $1.5$), similarity reaches its peak. Beyond this regime, the similarity begins to decline. This drop is caused by excessively strong residual injection, which introduces disruptive signals into the denoising dynamics and ultimately degrades structural coherence or visual quality. These patterns are clearly observed across all figures: \textit{Scene} (Fig.~\ref{fig:combined_scene}), \textit{Anime} (Fig.~\ref{fig:combined_anime}), \textit{People} (Fig.~\ref{fig:combined_people}), \textit{Style} (Fig.~\ref{fig:combined_style}), and \textit{Toy} (Fig.~\ref{fig:combined_toy}).

In contrast, the \emph{diversity} curves display a monotonic decreasing trend across all categories. As shown in the diversity plots of Fig.~\ref{fig:combined_scene}, Fig.~\ref{fig:combined_anime}, Fig.~\ref{fig:combined_people}, Fig.~\ref{fig:combined_style}, and Fig.~\ref{fig:combined_toy}, higher values of $\lambda$ constrain the sampling trajectory more strongly, reducing the degrees of freedom available during denoising. This reflects that the residual acts as a progressively stronger regularization signal—narrowing the generative distribution and enforcing more deterministic outputs that adhere to the target adaptation. The consistency of this effect across domains underscores the stabilizing influence of the delta signal on generation behavior.

Together, these results reveal a robust and interpretable similarity--diversity trade-off controlled by $\lambda$. Moderate values yield the best balance between fidelity and variety, whereas excessively strong guidance reduces both similarity and diversity due to over-dominance of the residual signal. The cross-domain consistency observed across Fig.~\ref{fig:combined_scene}--\ref{fig:combined_toy} demonstrates that DS behaves predictably and reliably under a unified mechanism across diverse visual categories.

\end{document}